\documentclass[preprint]{elsarticle}
\journal{Pattern Recognition}

\usepackage{amsmath}
\usepackage{graphicx}
\usepackage{booktabs}
\usepackage[capitalize]{cleveref}
\usepackage{subcaption}

\usepackage{wrapfig}
\usepackage{multirow}
\usepackage{amssymb}% http://ctan.org/pkg/amssymb
\usepackage{pifont}% http://ctan.org/pkg/pifont

\begin{document}

\begin{frontmatter}

%\title{Joint Depth and Flow Forecasting via Multimodal Recurrent Architectures}
\title{FLODCAST: Flow and Depth Forecasting via Multimodal Recurrent Architectures}

%\title{Forecasting Flow improves Depth Forecasting}

%\title{Joint Depth and Flow Forecast for Autonomous Driving}

\author[UNIFI]{Andrea Ciamarra}
\ead{andrea.ciamarra@unifi.it}
\author[UNISI]{Federico Becattini}
\ead{federico.becattini@unisi.it}
\author[UNIFI]{Lorenzo~Seidenari}
\ead{lorenzo.seidenari@unifi.it}
\author[UNIFI]{Alberto~Del~Bimbo}
\ead{alberto.delbimbo@unifi.it}

\address[UNIFI]{Dipartimento di Ingegneria dell'Informazione, University of Florence, Italy}
\address[UNISI]{Dipartimento di Ingegneria dell'Informazione e Scienze Matematiche, University of Siena, Italy}

\begin{abstract}
Forecasting motion and spatial positions of objects is of fundamental importance, especially in safety-critical settings such as autonomous driving. In this work, we address the issue by forecasting two different modalities that carry complementary information, namely optical flow and depth. To this end we propose FLODCAST a flow and depth forecasting model that leverages a multitask recurrent architecture, trained to jointly forecast both modalities at once. We stress the importance of training using flows and depth maps together, demonstrating that both tasks improve when the model is informed of the other modality. We train the proposed model to also perform predictions for several timesteps in the future. This provides better supervision and leads to more precise predictions, retaining the capability of the model to yield outputs autoregressively for any future time horizon. We test our model on the challenging Cityscapes dataset, obtaining state of the art results for both flow and depth forecasting. Thanks to the high quality of the generated flows, we also report benefits on the downstream task of segmentation forecasting, injecting our predictions in a flow-based mask-warping framework.
\end{abstract}

\begin{keyword}
depth forecasting \sep optical flow forecasting \sep segmentation
\end{keyword}

\end{frontmatter}

\section{Introduction}

Improving intelligent capabilities, in the context of robot navigation and autonomous agents, is fundamental to allow machines to better understand the observed scene and thus reason about it. These systems exploit sensors such as cameras or LiDARs to extract a visual signal from the environment in order to take action and interact with the world.
However, leveraging only the current frame to plan real-time decisions is challenging since dynamic scenes rapidly change over time.
Agents must understand how other objects are moving and must foresee possible dangerous outcomes of their decisions. A prominent direction with potential application in decision-making is to make predictions about future scenarios, which can also be used to detect upcoming events or behaviors in advance. 
%Anticipating the imminent future is a computer vision problem that involves an autonomous agent capable of understanding the surrounding context. This task is highly challenging, especially in urban scenarios, where multiple objects, like cars or pedestrians, can move freely in the environment.
This task is highly challenging in situations where multiple objects, like vehicles or people, can move freely in the environment.

The problem can be addressed from many angles, including understanding where agents will be in the near future, what actions they will take, how they will move, and how far they will be from a given observation point.
In practice, this translates into exploiting different features describing the scene or specific objects.
%In particular, several features of the scene can describe an urban environment.
For instance, road layout supports the agent in defining where to drive, while semantic segmentation contains pixel-level annotations of specific categories, e.g. road, buildings, cars or pedestrians, and gives a finer-grained knowledge of the scene. However, predictions may also regard future instance segmentations, allowing a machine to reason about single objects rather than category classes.
One way to summarize scene changes is to capture motion properties observed from a camera viewpoint. Optical flow is a dense field of displacement vectors and represents the pixel motion of adjacent frames \cite{zhai2021optical}.
Therefore, object motion can be incorporated in terms of 2D displacements using optical flow, even for future unobserved motion.
Nonetheless, in order to understand scene dynamics it is also considerable to predict depth maps to better identify objects in a 3D space. Such information can be estimated in advance for the near future and incorporated into a decision-making system that assists an autonomous agent to early plan the subsequent action to be taken.
Future prediction also involves information related to the surrounding environment. Therefore, this task can be accomplished by forecasting semantic segmentations \cite{luc2017predicting, terwilliger2019recurrent, saric2020warp}, which are connected to specific category classes, but also predicting future instance segmentations of moving objects \cite{luc2018predicting, sun2019predicting, hu2021apanet, lin2021predictive, ciamarra2022forecasting}, even considering optical flow predictions \cite{jin2017predicting, ciamarra2022forecasting}.

In summary, one can cast the forecasting problem from a high-level perspective, for instance forecasting semantic masks \cite{luc2018predicting, graber2021panoptic} or agent trajectories \cite{marchetti2022smemo, salzmann2020trajectron++}, as done in prior work. We instead choose to address the problem from a lower level, forecasting finer-grained information such as pixel-level optical flows and depth maps, which can then be leveraged to reason about high-level aspects such as forecasting semantic instances.
In this work, we focus on anticipating imminent future urban scenarios, by casting the problem in a multi-modal and multitasking approach, able to forecast both optical flows, which encode pixel displacements in the scene, and depth maps, which represent the estimated distance from the camera to the corresponding point in the image. 
Instead of anticipating the future for the next time step \cite{qi20193d, hu2020probabilistic} or in general for a single specific one \cite{nag2022far}, we propose to directly forecast multiple time steps ahead at a time, yet maintaining the model autoregressive to avoid the need of training timestep-specific models. Jointly forecasting depth and flow helps to achieve better performance in future predictions, thanks to information sharing across modalities. In addition, training with long-term supervision leads to smaller errors at inference time.
As a byproduct, we also leverage the recently proposed MaskNet \cite{ciamarra2022forecasting} to improve the downstream task of future instance segmentation in urban scenes with our predictions.

To summarize, our main contributions are the following:
\begin{itemize}
\item[(i)] We design a novel optical \underline{FLO}w and \underline{D}epth fore\underline{CAST}ing network (FLODCAST) that jointly estimates optical flow and depth for future frames autoregressively.
\item[(ii)] Our approach, which involves predicting multiple steps simultaneously, mitigates the accumulation of errors that typically impede the performance of autoregressive models. In this way, we preserve the autoregressive nature of the model, eliminating the need for training separate models for different time horizons.
\item[(iii)]Finally, FLODCAST achieves state-of-the-art performance in both optical flow and depth forecasting tasks, thereby emphasizing the necessity of jointly learning shared features from these two modalities.
\end{itemize}
%Moreover, we show that SceMCA, producing accurate optical flow and depth predictions, can be employed in accomplishing different future anticipation tasks. As a byproduct, we leverage a flow-based model, that has recently been proposed in \cite{ciamarra2022forecasting} to predict future instance segmentations of moving objects in urban scenes.
%The authors introduced MaskNet, a convolutional neural network that warps binary instances from the current frame onto the future one, through the guidance of the predicted flow field of the scene and the current semantic segmentation image.
%To generate future flows in advance, they also designed OFNet that produces optical flows by giving the past ones autoregressively. Considering that our SceMCA models more detailed features, coming from scene motion dynamics associated with the depth, we argue that our predicted flows can enhance MaskNet, especially for farthermost forecastings.

\section{Related work}

\paragraph{Depth Forecasting}

%Depth map estimation performed on the current frame does not help in foreseeing events or detecting future behaviours beforehand.
Several works have focused on learning to infer depth from monocular RGB cameras \cite{mertan2022single, eigen2015predicting, xie2020video}.
Nonetheless, relying on depth estimators on predicted future RGBs is hard, due to high uncertainty in predicting raw pixels \cite{ranzato2014video, mathieu2015deep, kalchbrenner2017video, van2017transformation, kwon2019predicting}. Therefore, other works propose to deal with depth anticipation for future frames, mostly known in the literature as depth forecasting or video depth forecasting.
%Therefore, we mainly focus on depth anticipation for future frames
%Rather than looking at the present and then making predictions on the current scenario, our aim is to reason about future unobserved frames, by providing crucial details of the scene and the environment in advance. Therefore, we focus on depth anticipation for future frames (mostly known in the literature as depth forecasting or video depth forecasting), since after-the-fact depth estimation can not help in foreseeing events or detecting future behaviours beforehand.
%Forecasting a depth map is harder than estimating it on the frame already available, due to uncertainty in appearance caused by object movement, occlusion and viewpoint.
Qi et. al \cite{qi20193d} introduce an entire framework for predicting 3D motion (both optical flow and depth map) and synthesizing the RGB with its semantic map for unobserved future frames. To this end, they leverage images, depth maps and semantic segmentations of past frames but they make predictions limited to the subsequent future frame, i.e. at the frame $t+1$.
Also limited to a single future timestep, Hu et. al \cite{hu2020probabilistic} design a probabilistic model for future video prediction, where scene features are learned from input images and are then used to build spatio-temporal representations, incorporating both local and global contexts. These features are finally fed into a recurrent model with separate decoders, each one forecasting semantic segmentation, depth and dense flow at the next future frame.
%Nag et. al \cite{nag2022far} propose a self-supervised method for depth estimation for unobserved frames directly after k time instants. 
Nag et. al \cite{nag2022far} propose a self-supervised method for depth estimation directly at the k-th frame after the last observed one, i.e. at $t+k$. %after k time instants. 
By means of a feature forecasting module, they learn to map pyramid features extracted from past sequences of both RGBs and optical flows to future features, exploiting a series of ConvGRUs and ConvLSTMs for spatio-temporal relationships in the past.
%They train a pose estimation module conditioned to the future frame $I_{t+k}$, i.e. passing $\{I_{t}, I_{t+2k}\}$ as source images instead of the two consecutive frames $\{I_{t-1}, I_{t+1}\}$, to be used as supervised signal in order to reconstruct the predicted depth through the future view.
With the same goal, Boulahbal et. al \cite{boulahbal2022forecasting} design an end-to-end self-supervised approach by using a hybrid model based on CNN and Transformer that predicts depth map and ego-motion at $t+k$ by processing an input sequence of past frames.
Differently from prior work, we predict both dense optical flows and depth maps, also leveraging both modalities as inputs. We directly predict several timesteps ahead simultaneously while retaining autoregressive capabilities, that allows the model to accurately predict far into the future.

\paragraph{Flow Forecasting}

%Optical flow is the task of estimating the per-pixel motion between video frames.
%, by exploiting encoder-decoder, spatial pyramid network, iterative approaches
Optical flow estimation has been largely studied in the past \cite{tu2019survey, zhai2021optical}. Consolidated deep learning approaches have addressed this problem with promising results \cite{ilg2017flownet, sun2018pwc, teed2020raft}, also exploiting transformer-based architectures \cite{huang2022flowformer, shi2023flowformer++, lu2023transflow}. However, these methods are designed to estimate the optical flow by accessing adjacent frames as they are available to the network.
Different approaches have been introduced incorporating optical flow features to infer imminent future scenarios under different points of view, such as predicting depth maps \cite{nag2022far}, semantic segmentations \cite{terwilliger2019recurrent, saric2020warp} and instance segmentations \cite{ciamarra2022forecasting}. Multitasking methods also exist \cite{jin2017predicting, yin2018geonet, qi20193d}.

Many works leverage motion features for future predictions to perform several specific tasks, ranging from semantic segmentation \cite{jin2017predicting, luc2017predicting, terwilliger2019recurrent, saric2020warp}, instance-level segmentation \cite{ciamarra2022forecasting} and depth estimation \cite{qi20193d, hu2020probabilistic, nag2022far}.
However, just a few approaches have specifically addressed the task of optical flow forecasting, i.e. the problem of anticipating the optical flow for future scenes. Jin et. al \cite{jin2017predicting} was the first to propose a framework, which jointly predicted optical flow and semantic segmentation for the next frame using the past ones. To make predictions for multiple time steps, they just iterate a two-step finetuned model so to alleviate the propagation error. Ciamarra et. al \cite{ciamarra2022forecasting} instead introduced OFNet, a recurrent model able to predict the optical flow for the next time step exploiting spatio-temporal features from a ConvLSTM. Such features are learned to generate a sequence of optical flows shifted by one time step ahead from the input sequence. Without finetuning, the recurrent nature of the model allows OFNet to make predictions for any time steps ahead. %, each time by taking the last prediction in the current output sequence and appending it to the input.
Considering the high uncertainty of the future, all the proposed methods \cite{terwilliger2019recurrent, jin2017predicting, yin2018geonet, qi20193d, ciamarra2022forecasting} are typically trained to make predictions at the single time step ahead, and then used for the future ones by autoregressively providing in input the predictions obtained at the previous iterations.
We, instead, address a more general forecasting task, with the purpose of providing future optical flows directly for multiple time steps ahead, by exploiting both past flows and the corresponding depth maps. We also make use of depth maps as input because our framework is designed as a novel multitask and multimodal approach to also generate future depth maps. 

To the best of our knowledge, we are the first to jointly forecast optical flows and depth maps for multiple consecutive frames into the future. Besides, we do not require other information (even during training), like camera pose estimation, which is usually needed to deal with monocular depth estimation.

\section{Method}

%Inspired by \cite{nag2022far}, we propose to address the depth forecasting task, by formulating a more general problem for depth-and-flow anticipation. 
In this work we introduce FLODCAST, a novel approach for predicting optical flow and depth map jointly for future unobserved frames from an ego-vehicle perspective applied to autonomous driving context.
%This section is organized as follows: first, we will describe the problem formulation as a forecasting task in the paragraph \ref{sec:problemdef}, then we will present our multitask and multimodal approach for forecasting optical flow and depth map in the paragraph \ref{sec:flodcast}. Finally, we discuss the loss functions involved to train our model in the paragraph \ref{sec:loss}.  

\subsection{Problem Definition}\label{sec:problemdef}
Given a sequence $\mathbf{S}=\{I_t\}$ of frames, let $\mathbf{D}=\{D_1, D_2, \ldots, D_T\}$ be the depth map sequence extracted from the last T frames of $\mathbf{S}$. Likewise, we define $\mathbf{OF}=\{OF_1, OF_2, \ldots, OF_{T}\}$ the corresponding optical flows computed every two consecutive frames in $\mathbf{S}$, such that $OF_{t}=\textit{Flow}(I_{t-1}, I_{t})$, with $t \in \left[ 1,T \right]$, encodes the motion of the source frame $I_{t-1}$ onto the target frame $I_{t}$. Our purpose is to anticipate flow and depth maps for future frames after $K$ time instants, i.e. forecasting $D_{T+K}$ and $OF_{T+k}$ for the frame $I_{T+K}$.
%\footnote{More precisely, for the same frame $I_{j+1}$ in a video sequence $\mathbf{S}$, we denote the pair $(OF_j, D_j)$, where the flow field $OF_{j}=\textit{Flow}(I_j, I_{j+1})$ and depth is $D_{j+1}$.}

%\todo{\textbf{FORSE UTILE DA SCRIVERE DA QUALCHE PARTE?} The importance of anticipating flow and depth for future scenarios is two-fold in navigation systems: such an agent (i) can exploit scene motion (according to its camera) and the depth, both estimated in advance, in order to plan the next action to be taken, and (ii) being smoother in autonomous driving, as the distance and motion estimated in the whole scene are the two the most useful information to already have on-hand.}

The importance of jointly anticipating flow and depth stems from the nature of the two modalities. Optical flow is a two-dimensional projection of the three-dimensional motion of the world onto the image plane \cite{vedula1999three}. An object in the foreground moving fast produces a large displacement, whereas when it comes far from the observer, moving at the same speed, it generates a very small displacement. Therefore, knowledge about the depth of such an object can help to model its future dynamics. Vice-versa, observing the motion of an object can provide information about its distance from the camera.
Overall, by jointly modeling optical flow and depth we can represent the 3D scene displacement at time $t$ in terms of the components $(u, v, d, t)$, where $(u, v)$ are the horizontal and vertical components of $OF_t$ and $d$ is the depth map.

\begin{figure}[t]
\centering
\includegraphics[width=\linewidth]{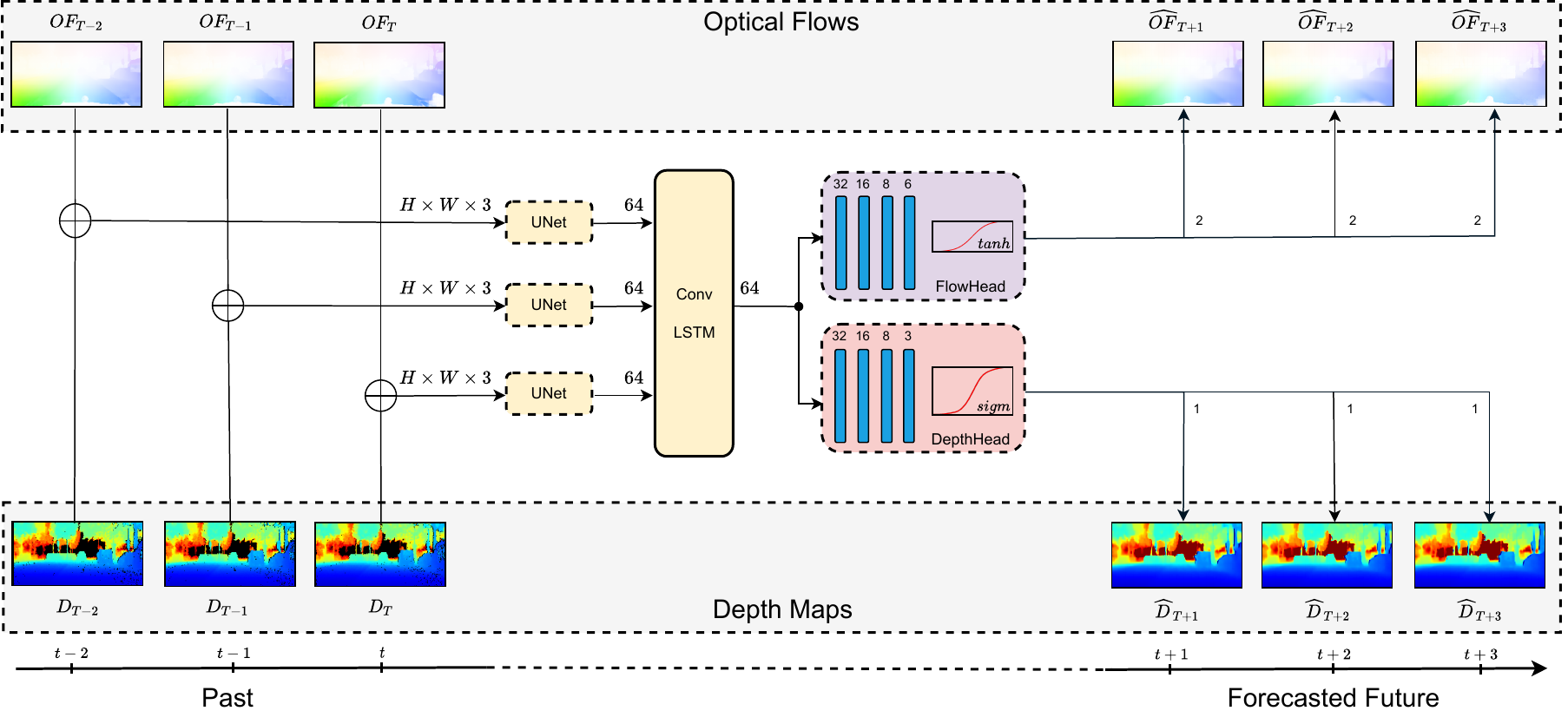}
\caption{FLODCAST forecasts both future flows and depth maps from the past ones autoregressively. For each time step, we aggregate flow and depth at the last channel (by the concatenation operator, $\oplus$), then 64-channel features are extracted through a UNet \cite{ronneberger2015u} backbone. Finally, predictions are obtained from two dedicated fully convolutional heads.}
\label{fig:scemca}
\end{figure}

%\subsection{Scene Motion Forecasting}
%\subsection{Joint Depth and Optical Flow Forecasting}
%\subsection{Joint Flow and Depth Forecasting via Multimodal Recurrent Architectures}
\subsection{Flow and Depth Forecasting via Multimodal Recurrent Architectures}\label{sec:flodcast}

%The optical flow predictor OFNet in \cite{ciamarra2022forecasting} was designed to be used autoregressively at any future time step, in order to be employed in a framework addressing the future instance segmentation task of moving objects. This network anticipates the dense flow estimated for the following time step, after observing $T=6$ optical flows in the past, i.e. OFNet predicts $\widehat{OF}_7$ through $OF_{1:6}=\{OF_1, OF_2, \ldots, OF_6\}$. Considering OFNet recursively produces optical flows in the future, by appending each time the last prediction at the end of the input sequence, it has three main drawbacks: (i) it takes time to make predictions even in the near future, while (ii) accumulating errors for not so long temporal horizons, e.g. after $t+9$ time instants ahead (named mid-term predictions), and (iii) estimating motions of objects being at different scales through the optical flow as a unique source of information, may produce bad predictions since they can move at different speeds.

%Motivated by these issues,
We design \textbf{FLODCAST}, a novel optical \textbf{\underline{FLO}}w and \textbf{\underline{D}}epth fore\textbf{\underline{CAST}}ing network that anticipates both modalities at each future time step by observing the past ones. An overview of FLODCAST is shown in Fig. \ref{fig:scemca}.

FLODCAST takes a sequence $X=\{X_1, X_2, \ldots, X_T\}$ of $T$ past observations composed of dense optical flows and depth maps. 
%FLODCAST takes a sequence of $T$ dense dynamic motion-and-geometric feature (MGs) in the past $\{MG_1, MG_2, \ldots, MG_T\}$ and directly produces two sequences each long $T$, which encode the future, i.e. $\{\widehat{OF}_{T+1}, \widehat{OF}_{T+2}, \ldots, \widehat{OF}_{2T}\}$ and $\{\widehat{D}_{T+1}, \widehat{D}_{T+2}, \ldots, \widehat{D}_{2T}\}$.
In detail, each $X_t$ encodes the input features for the image $I_t$ in the past, that are obtained by concatenating the optical flow $OF_t$ with the depth map $D_t$. In other words, $X_t = \left( OF_t \oplus D_t\right)$.
The model generates as output a sequence $\widehat{X}=\{\widehat{X}_{T+1}, \widehat{X}_{T+2}, \ldots, \widehat{X}_{T+K}\}$, that is a sequence of $K$ future optical flows and $K$ depth maps.
We set $T=3$ and $K=3$ in all our experiments.

%As we will discuss later in the results (see Section \ref{sec:result}), the third dimension, which was missing in the OFNet implementation, will drastically reduce the prediction errors at long temporal horizons while anticipating not only the future optical flow but also the depth map for unobserved frames, thus making FLODCAST be a more complete approach capable of understanding the environment from an ego-vehicle perspective.%, including the depth forecasting problem for future frames.

%In order to allow the model to handle multi-modality predictions, we append two convolutional branches after extracting input feature from $\{MG_k\}_{k=1}^T$, and each set of weights is trained end-to-end to generate a temporal sequence of future optical flows or depth maps. 
Since optical flows and depth maps encode very different information about the scene, we add two separate heads after extracting features from the input in order to handle multimodal predictions. Therefore, we feed in input a sequence of concatenated optical flows and depths $\{X_1, X_2, \ldots, X_T\}$ to a recurrent ConvLSTM network, in which a UNet backbone is used to extract features at 64 channels for each input $X_t$, $t=1, \ldots, T$, so to output a tensor of size $(H \times W \times 64)$, where $(H \times W)$ is the input resolution. Our feature extractor is the same UNet architecture as in \cite{ciamarra2022forecasting}, i.e. a fully convolutional encoder-decoder network with skip connections, consisting of 5 layers with filters $\{64, 128, 256, 512, 1024\}$ respectively. These 64-channel features capture meaningful spatio-temporal contexts of the input representation. The features are then passed to the two convolutional heads, which are end-to-end trained to simultaneously generate the sequence of future optical flows and depth maps (respectively depicted by the purple and the red blocks in the right side of Fig. \ref{fig:scemca}). Each head is a fully convolutional network made of sequences of Conv2D+ReLUs with $\{32, 16, 8\}$ filters. %The aim of having separated these two additional networks is to empower the prediction capabilities so to make the model prone to generate specialized feature for optical flow and depth, that are complementary each other. 
Finally, we append at the end of the optical flow head a convolution operation with $2 \times K$ channels and we use a $tanh$ activation function, so to produce the $(u,v)$ flow field values normalized in $(-1, 1)$. Instead, after the depth head, we attach a convolution operation with a $K$ channels and a sigmoid activation in order to get depth maps normalized in $(0, 1)$.
Instead of outputting one prediction at a time as in prior work \cite{ciamarra2022forecasting}, we directly generate $K$ flows and depth maps simultaneously, to make the model faster compared to autoregressive models which would require looping over future steps.

\subsection{Loss}\label{sec:loss}

To train FLODCAST we compute a linear transformation of the original input values, by rescaling depth map values in $[0,1]$ and optical flows in $[-1,1]$ through a min-max normalization, with minimum and maximum values computed over the training set.
%Inspired by \cite{laina2016deeper}, we use the reverse Huber loss, called \textit{BerHu}, since we found in our experiments that it is more effective than other existing regression losses, e.g. Huber loss, MSE, MAE.
Inspired by \cite{laina2016deeper}, we use the reverse Huber loss, called \textit{BerHu} for two main reasons: (i) it has a good balance between the two L1 and L2 norms since it puts high weight towards values with a high residual, while being sensitive for small errors; (ii) it is also proved to be more appropriate in case of heavy-tailed distributions~\cite{laina2016deeper}, that perfectly suits our depth distribution, as shown in Fig. \ref{fig:depthhisttrain}. BerHu minimizes the prediction error, through either the L2 or L1 loss according to a specific threshold $c$ calculated for each batch during the training stage. Let $x=\hat{y}-y$ be the difference between the prediction and the corresponding ground truth. This loss $\mathcal{B}(x)$ is formally defined as:
\begin{equation}
\mathcal{B}(x) = 
\begin{cases}
|x|, & |x| \leq |c|\\[5pt]
\frac{x^2+c^2}{2c}, & \text{otherwise}
\end{cases}
\label{eq:berhufunction}
\end{equation}
Thus, we formulate our compound loss, using a linear combination of the optical flow loss $\mathcal{L}_{\text{flow}}$ and the depth loss $\mathcal{L}_{\text{depth}}$ (Eq. \ref{eq:lossfunction}):
\begin{equation}
\mathcal{L} = \alpha \, \mathcal{L}_{\text{flow}} + \beta \, \mathcal{L}_{\text{depth}}
\label{eq:lossfunction}
\end{equation}
Specifically, we apply the reverse Huber loss to minimize both the optical flow and depth predictions, using the same loss formulation, since the threshold $c$ is computed for each modality, and that value depends on the current batch data. Therefore, $\mathcal{L}_{\text{flow}}$ is the loss function for the optical flow computed as: 
\begin{equation}
\mathcal{L}_{\text{flow}} = \frac{1}{M} \sum_{j=1}^{M} \mathcal{B}(|OF_{j} - \widehat{OF}_j|)
\label{eq:flowlossfunction}
\end{equation}
where $M=B \times R \times 2$, since the flow field has $(u,v)$ components over $R$ image pixels and $B$ is the batch size, whereas $OF_j$ and $\widehat{OF}_j$ are the optical flows, respectively of the ground truth and the prediction at the pixel $j$. Likewise, we do the same for the depth loss $\mathcal{L}_{\text{depth}}$:
\begin{equation}
\mathcal{L}_{\text{depth}} = \frac{1}{P} \sum_{j=1}^{P} \mathcal{B}(|D_{j} - \widehat{D}_j|)
\label{eq:depthlossfunction}
\end{equation}
where $P=B \times R$, $D_j$ and $\widehat{D}_j$ are the depth maps, respectively of the ground truth and the prediction at the pixel $j$. We follow~\cite{laina2016deeper} and we set $c=\frac{1}{5}max_j(|y_j-\hat{y}_j|)$, i.e. the 20\% of the maximum absolute error between predictions and ground truth in the current batch over all pixels.

\section{Results}\label{sec:result}

% In this section, we report experimental results on Cityscapes \cite{cordts2016cityscapes} for the depth and flow forecasting tasks. Specifically, we will first describe the dataset in Section \ref{sec:cityscapesdataset}. Then, in Section \ref{sec:experimentalsetting}, we will provide further details to be taken into account on addressing the depth forecasting problem in this dataset, even reporting our experimental settings. We listed in Section \ref{sec:metrics} all the metrics used to evaluate performance and accuracy and results are shown in Section \ref{sec:futdepthest}. We conduct an ablation study in Section \ref{sec:ablationstudy} that supports our choices for FLODCAST. In addition, in Section \ref{sec:performance} we discuss the performance details of FLODCAST at runtime.
% Besides, in Section \ref{sec:videoprediction} we will also evaluate segmentation forecasting results in terms of both instance and semantic segmentations on future unobserved frames, by exploiting our optical flows forecasted by FLODCAST and the warping neural network MaskNet designed in~\cite{ciamarra2022forecasting}.

In this section we report our forecasting results on Cityscapes \cite{cordts2016cityscapes} for the depth and flow forecasting tasks. We first describe the experimental setting and the metrics used to evaluate our approach. Then, we present our results, comparing FLODCAST to state-of-the-art approaches. We also present ablation studies to better highlight the importance of all the modules in the architecture. Besides, in Sec. \ref{sec:videoprediction}, we show that our approach can be easily applied to downstream tasks such as semantic segmentation and instance segmentation forecasting, demonstrating improvements, especially at farther prediction horizons.

\subsection{Dataset}\label{sec:cityscapesdataset}

For evaluation, we use Cityscapes~\cite{cordts2016cityscapes}, which is a large urban dataset with very challenging dynamics, recorded in several German cities. Each sequence consists of 30 frames at a resolution of $1024 \times 2048$. Cityscapes contains 5000 sequences, split in 2975 for train, 500 for validation and 1525 for testing. Different annotations are available. In particular, we leverage precomputed disparity maps for all frames, from which depth maps can be extracted through the camera parameters. There are also both instance and semantic segmentations that are available at the 20-th frame of each sequence.

\begin{figure}[t]
\includegraphics[width=\linewidth]{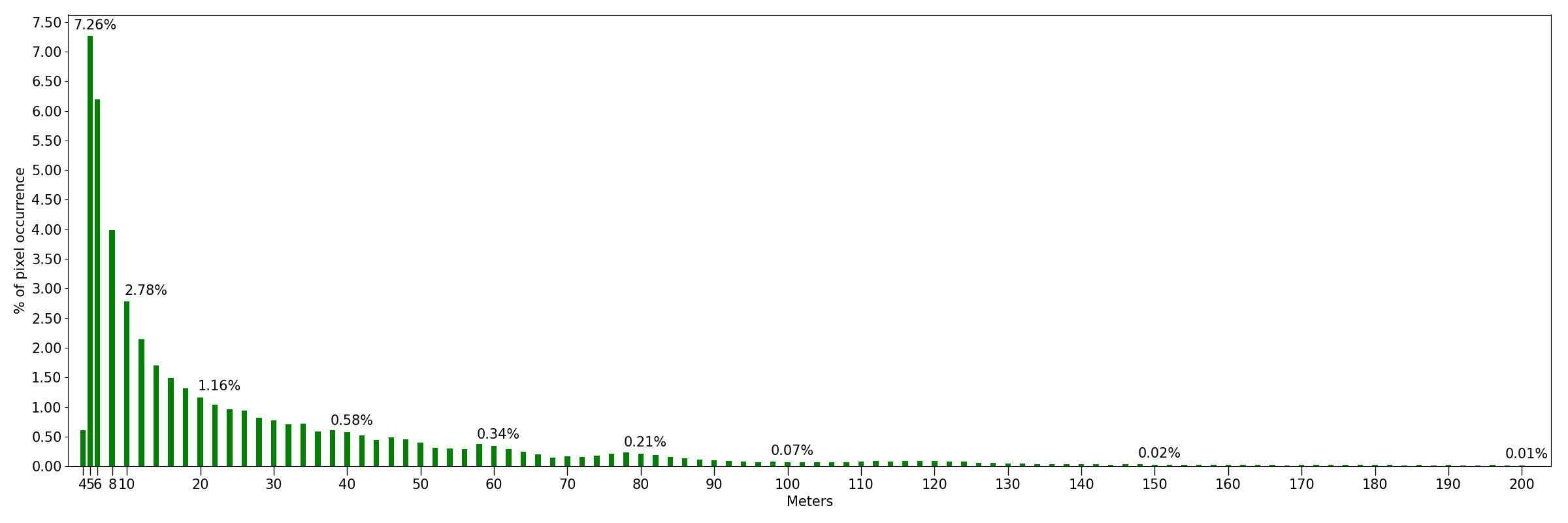}
\caption{Distribution of depth values grouped by distance on the Cityscapes training set. Note that depth values below 3 meters are not present in the dataset.}\label{fig:depthhisttrain}
\end{figure}

    \subsection{Experimental setting}\label{sec:experimentalsetting}

%We compare our method with previous works for depth and flow forecasting tasks.
%We compute optical flows at every pair of adjacent frames for each video sequence in the whole dataset, through FlowNet-c with pretrained weights from FlowNet2 \cite{ilg2017flownet}.
We compute optical flows using FLowNet2 \cite{ilg2017flownet} (pretrained FlowNet2-c) and rescale them according to the maximum and minimum values in the training set, so to have normalized values in $(-1,1)$. Depth maps $D$ are obtained using disparity data $d$ and camera parameters (focal length $f$ and baseline $b$), i.e. by computing $D = f \cdot b / d$. Invalid measurements or zero-disparity values are set to 0. 
To normalize depth maps, we observe that most depth values fall within 150m in the training set (Fig. \ref{fig:depthhisttrain}). Thus, we cap values at 150m and then normalize them in $(0,1)$.
All frames are rescaled at $128 \times 256$ px for both data sources to accelerate learning.
We train FLODCAST for 30 epochs using Adam and learning rate 0.0001.
%We set $T=3$, so to generate future sequences of both optical flows and depth maps, for the following 3 time steps, instead of outputting one prediction at a time \cite{ciamarra2022forecasting}.
To balance the two losses in Eq. \ref{eq:lossfunction}, we set $\alpha=10$ and $\beta = 1$. At inference time we recursively employ the model by feeding as input previous predictions to reach farther time horizons. We provide outputs at a resolution of $256 \times 512$, following \cite{pilzer2018unsupervised}, by doubling the resolution.
FLODCAST has approximately 31.4M trainable parameters. The whole training takes 58 hours on a single GPU NVIDIA Titan RTX with 24GB using a batch size of 12.

\subsection{Evaluation metrics}\label{sec:metrics}

We quantitatively evaluate depth forecasting using standard metrics as in \cite{eigen2014depth}: (i) absolute relative difference (AbsRel), (ii) squared relative difference (SqRel), (iii) root mean squared error (RMSE) and (iv) logarithmic scale-invariant RMSE (RMSE-Log), defined as follows:
\\
\noindent
\begin{tabular}{@{}p{.4\linewidth}@{}p{.02\linewidth}@{}p{.58\linewidth}@{}}
\begin{equation}\label{equ:depthmetricone}
\text{AbsRel} = \frac{1}{N} \sum_{i=1}^{N} \frac{|y_i-\hat{y}_i|}{y_i} 
\end{equation} & &
\begin{equation}\label{equ:depthmetricthree}
\text{RMSE} = \sqrt{\frac{1}{N} \sum_{i=1}^{N} |y_i-\hat{y}_i|^2}
\end{equation} \\
\vspace*{-\baselineskip}
\begin{equation}\label{equ:depthmetrictwo}
\text{SqRel} = \frac{1}{N} \sum_{i=1}^{N} \frac{(y_i-\hat{y}_i)^2}{y_i}
\end{equation} & & 
\vspace*{-\baselineskip}
\begin{equation}\label{equ:depthmetricfour}
\text{RMSE-Log} = \frac{1}{N} \sum_{i=1}^{N} d_i^2 - \frac{1}{N^2} \left( \sum_{i=1}^{N} d_i \right)^2
\end{equation}
\end{tabular}

where $y$ and $\hat{y}$ are the ground truth and the prediction, each with $N$ pixels indexed by $i$, while $d=\log \hat{y} - \log y$ is their difference in logarithmic scale.
AbsRel and SqRel are errors that can be also calculated at pixel-level, instead RMSE, RMSE-Log measure mistakes averaged on the whole image.
In particular, AbsRel draws attention to the absolute difference between the prediction and the target with respect to the ground truth itself (e.g. an AbsRel of 0.1 means that the error is 10\% of the ground truth), which makes it suitable for a fine-grained understanding.
The SqRel instead emphasizes large errors since the difference is squared.
RMSE is the root of the mean squared errors while RMSE-Log, introduced in \cite{eigen2014depth}, is an L2 loss with a negative term used to keep relative depth relations between all image pixels, i.e an imperfect prediction will have lower error when its mistakes are consistent with one another.
%DA RIVEDERE [https://medium.com/@omarbarakat1995/depth-estimation-with-deep-neural-networks-part-1-5fa6d2237d0d]

We also measure the percentage of inliers with different thresholds \cite{eigen2014depth}, i.e. the percentage of predicted values $\hat{y}_i$ for which the ratio $\delta$ with the ground truth $y_i$ is lower than a threshold $\tau$: 
%$\delta<\tau$:
%
\begin{equation}
\text{\% of } \hat{y} \; \text{ s.t. } \; \max \left( \frac{y_i}{\hat{y}_i}, \frac{\hat{y}_i}{y_i} \right) = \delta < \tau 
\label{eq:acceq}
\end{equation}
with $\tau=\{1.25,\, 1.25^2, \, 1.25^3\}$.\\\\
We assess the performance of the flow forecasting task, by computing the mean squared error between the prediction and the groundtruth on both the two flow channels, using Eq. \ref{eq:mseflow}, and averaging them, as done in \cite{ciamarra2022forecasting}:
\begin{equation}
\text{MSE}_c = \frac{1}{H \: W} \sum_{i=1}^{H} \sum_{j=1}^{W} \left( f_{c}(i,j) - {\widehat{f}}_{c}(i,j) \right) ^2
\label{eq:mseflow}
\end{equation}
where $\text{MSE}_c$ is the error referred to the channel $c:=\{u,v\}$ between the ground truth optical flow field $f_{c}(i,j)$ and the prediction $\widehat{f}_{c}(i,j)$ at the pixel $(i,j)$ and H and W is height and width respectively. We also report the average end-point-error EPE \cite{baker2011database}, which measures the per-pixel euclidean distance between the prediction and the ground truth averaged among all the image pixels:
\begin{equation}
\text{EPE} = \frac{1}{H \: W} \sum_{i=1}^{H \: W} \sqrt{(\hat{u_i}-u_i)^2 + (\hat{v_i}-v_i)^2}
\label{eq:epeflow}
\end{equation}
where $(u_i, v_i)$ are the horizontal and vertical components of the optical flow ground truth, likewise $(\hat{u_i}, \hat{v_i})$ are the corresponding components of the prediction, at the $i-th$ pixel.%\\

%Instead, for the segmentation forecasting task we use the mAP and mAP50 metrics for instance segmentation, and mIoU (mean IoU) for semantic segmentation, evaluated on 8 different categories of moving objects, as done in previous works \cite{luc2018predicting, hu2021apanet, ciamarra2022forecasting}. The metric mAP50 is the mean average precision that counts an instance as correct if matching its groundtruth for at least 50\% of intersection-over-union (IoU). Instead, mAP is obtained by averaged 10 equally spaced IoU thresholds from 50\% to 95\%. 

\subsection{Future Depth Estimation}\label{sec:futdepthest}

We evaluate our approach for future depth estimation on Cityscapes.
%In this dataset, several methods have addressed depth estimation alone \cite{pilzer2018unsupervised, casser2019unsupervised}, or in combination with other tasks \cite{wang2020sdc, gao2022panopticdepth}. Rather, other works \cite{hu2020probabilistic, nag2022far} explore depth forecasting, by anticipating depth maps for future unseen frames.
As in prior works, e.g. \cite{hu2020probabilistic}, we evaluate our method after $t+k$ frames, both at short-term ($k=5$, after 0.29 sec) and at mid-term ($k=10$, after 0.59 sec).
% Tutta questa roba degli indici dei frame secondo me genera un sacco di confusione, per ora eviterei di metterla
%while keep following the same evaluation strategy as done for semantic forecasting in \cite{luc2017predicting, luc2018predicting}, i.e. by comparing errors and metrics between the prediction and the ground truth at the same 20-th frame of each video sequence.
%We use FLODCAST autoregressively, by feeding $\{OF_{13}, OF_{14}, OF_{15}\}$ and $\{D_{13}, D_{14}, D_{15}\}$ for short-term, and $\{OF_{8}, OF_{9}, OF_{10}\}$ and $\{D_{8}, D_{9}, D_{10}\}$ for mid-term.

Since there is no official evaluation protocol for depth forecasting on Cityscapes and considering the statistics in the training set (see Fig. \ref{fig:depthhisttrain}), in which pixel occurrences strongly decrease as the depth increase, we clip values at 80 meters as done in prior work for depth estimation \cite{pilzer2018unsupervised, casser2019unsupervised}.

%
% \begin{figure}[t]
%\includegraphics[width=\linewidth]{images/occ_hist_test.png}
%\caption{Depth map occurrence distribution on Cityscapes train set. Only a small amount of pixels are far (around 0.4\% from 80 meters, as shown in the last bin).}\label{fig:occhisttest}
%\end{figure}
%

%Additionally, we analyze the depth map occurrences on the test set and depicted in Fig. \ref{fig:occhisttest}, we found that after 50 meters the percentage of pixel occurrences drastically decreases. Moreover, few pixels are far away, specifically depth values far at least 80 meters are less than 0.4\% of the test set. This motivated us to cap depth maps to 80 meters (as actually done in \cite{casser2019unsupervised}).
%Considering our analysis, we hope that future works will keep following this evaluation protocol for depth estimation on Cityscapes dataset.
For our experiments, we evaluate predictions using the same protocol of \cite{pilzer2018unsupervised}, i.e. by cropping out the bottom 20\% of the image to remove the car hood, which is visible in every frame, then we rescale the frames at $256 \times 512$. In addition, we mask out ground truth pixels that are farther than the 80m threshold.
%Therefore, we compare ground-truth pixels with our predictions generated by FLODCAST, which are obtained by denormalizing depth maps and clipping at 80 meters.
%Since FLODCAST outputs normalized depth predictions we get values back by first multiplying by the maximum value so to get values in $(0, 150)$, then we clip at 80 meters.

We compare our approach with existing methods \cite{qi20193d, hu2020probabilistic, nag2022far}. We also consider the depth estimation method of \cite{godard2019digging}, which is adapted to depth forecasting through a multi-scale F2F~\cite{luc2018predicting} before the decoder, and the future instance segmentation model~\cite{sun2019predicting} adapted to generate future depth estimation of the predicted features, as previously done in \cite{nag2022far}. We also report the trivial \textit{Copy last} baseline~\cite{nag2022far}, as a lower bound.
Quantitative results for depth forecasting are reported in Table \ref{table:depthresults}.

\begin{table}[t]
\centering
\caption{Quantitative results for depth forecasting after $t+k$ on Cityscapes test set, both at short-term and mid-term predictions, i.e. at $k=5$ and $k=10$ respectively.}\label{table:depthresults}
\resizebox{0.93\linewidth}{!}{
%\begin{minipage}{\textwidth}
\begin{tabular}{@{}c|cccc|ccc@{}}
\toprule
\multicolumn{8}{c}{Short term $k=5$} \\ \toprule
& \multicolumn{4}{c|}{Lower is better $\downarrow$} & \multicolumn{3}{c}{Higher is better $\uparrow$} \\ \toprule
Method & AbsRel & SqRel & RMSE & RMSE-Log & $\delta < 1.25$ & $\delta < 1.25^2$ & $\delta < 1.25^3$ \\ %\midrule
Copy last & 0.257 & 4.238 & 7.273 & 0.448 & 0.765 & 0.893 & 0.940 \\ \midrule
Qi et al.~\cite{qi20193d} & 0.208 & 1.768 & 6.865 & 0.283 & 0.678 & 0.885 & 0.957 \\
Hu et al.~\cite{hu2020probabilistic} & 0.182 & 1.481 & 6.501 & 0.267 & 0.725 & 0.906 & 0.963 \\
Sun et al.~\cite{sun2019predicting} & 0.227 & 3.800 & 6.910 & 0.414 & 0.801 & 0.913 & 0.950 \\ %\midrule
Goddard et al.~\cite{godard2019digging} & 0.193 & 1.438 & 5.887 & 0.234 & 0.836 & 0.930 & 0.958 \\ 
DeFNet~\cite{nag2022far} & 0.174 & 1.296 & 5.857 & 0.233 & 0.793 & 0.931 & 0.973 \\ \midrule
%FLODCAST w/o flow (OLD) & \underline{0.103} & 1.338 & 6.170 & \textbf{0.181} & \underline{0.900} & \underline{0.957} & \underline{0.977} \\
%SceMCA w/o flow & 0.103 & 1.338 & 6.170 & \textbf{0.034} & 0.900 & 0.957 & 0.977 \\
%SceMCA (Ours) &  \textbf{0.082} & 0.985 & 5.465 & 0.548 & \textbf{0.937} & \textbf{0.970} & \textbf{0.980} \\
%FLODCAST (OLD) & \textbf{0.086} & \textbf{0.932} & \textbf{5.347} & \underline{0.193} & \textbf{0.921} & \textbf{0.962} & \textbf{0.980} \\\midrule
FLOODCAST w/o flow & \underline{0.084} & \underline{1.081} & \underline{5.536} & \underline{0.196} & \underline{0.920} & \underline{0.963} & \underline{0.980} \\
\textbf{FLOODCAST} & \textbf{0.074} & \textbf{0.843} & \textbf{4.965} & \textbf{0.169} & \textbf{0.936} & \textbf{0.971} & \textbf{0.984}
\\\midrule

\multicolumn{8}{c}{Mid term $k=10$} \\ \toprule
& \multicolumn{4}{c|}{Lower is better $\downarrow$} & \multicolumn{3}{c}{Higher is better $\uparrow$} \\ \toprule
Method & AbsRel & SqRel & RMSE & RMSE-Log & $\delta < 1.25$ & $\delta < 1.25^2$ & $\delta < 1.25^3$ \\ %\midrule
Copy last & 0.304 & 5.006 & 8.319 & 0.517 & 0.511 & 0.781 & 0.802 \\ \midrule
Qi et al.~\cite{qi20193d} & 0.224 & 3.015 & 7.661 & 0.394 & 0.718 & 0.857 & 0.881 \\
Hu et al.~\cite{hu2020probabilistic} & 0.195 & \underline{1.712} & \textbf{6.375} & 0.299 & 0.735 & 0.896 & 0.928\\
Sun et al.~\cite{sun2019predicting} & 0.259 & 4.115 & 7.842 & 0.428 & 0.695 & 0.817 & 0.842 \\ %\midrule
Goddard et al.~\cite{godard2019digging} & 0.211 & 2.478 & 7.266 & 0.357 & 0.724 & 0.853 & 0.882 \\
DeFNet~\cite{nag2022far} & 0.192 & 1.719 & \underline{6.388} & 0.298 & 0.742 & 0.900 & 0.927 \\ \midrule
%FLODCAST w/o flow (OLD) & \underline{0.164} & 2.924 & 8.407 & \textbf{0.290} & \underline{0.827} & \underline{0.912} & \underline{0.948} \\
%SceMCA w/o flow & 0.164 & 2.924 & 8.407 & \textbf{0.054} & 0.827 & 0.912 & 0.948 \\
%SceMCA (Ours) & \textbf{0.120} & 1.701 & 7.259 & 0.736 & \textbf{0.880} & \textbf{0.943} & \textbf{0.964} \\
%FLODCAST (OLD) & \textbf{0.122} & \textbf{1.617} & 7.094 & \underline{0.296} & \textbf{0.867} & \textbf{0.936} & \textbf{0.964} \\ \bottomrule
FLOODCAST w/o flow & \underline{0.130} & 2.103 & 7.525 & 0.320 & \underline{0.863} & \underline{0.931} & \underline{0.959} \\
\textbf{FLODCAST} & \textbf{0.112} & \textbf{1.593} & 6.638 & \textbf{0.231} & \textbf{0.891} & \textbf{0.947} & \textbf{0.969} \\ \bottomrule
\end{tabular}
%\end{minipage}
}
\end{table}
%

%We exceed all the previous methods at short-term predictions.
%Specifically, we obtained relevant improvements with respect to all previous supervised approaches, especially in terms of AbsRel and SqRel (e.g. +52\% and +37\% respectively from Hu et. al) as well as the recent unsupervised method DeFNet (+50\% and +28\%), that employs both RGB frames and optical flows, even considering the pose estimation during the training. Instead, we exploit depth maps and optical flows as sources of information, since together provide scene motion features through time by means of a recurrent network. We believe that FLODCAST is capable of detecting such clues by extrapolating scene motion from a past sequence, 
We exceed all the previous methods at short-term and mid-term predictions. Specifically, we beat all the existing approaches at short-term by a large margin for all the metrics, also reporting the highest inlier percentage. 
At mid-term term we exceed all the state-of-the-art approaches, in terms of AbsRel and SqRel, including the recent DeFNet (-42\% and -8\%), which employs both RGB frames and optical flows, even considering the camera pose during the training.
%(e.g. -43\% and -7\% respectively from the Hu et. al)
%as well as the recent unsupervised method DeFNet (-42\% and -8\%),
Differently from DeFNet, we exploit depth maps and optical flows as sources of information, since they provide complementary features related to motion and geometric structure of the scene by means of a recurrent network. We believe that FLODCAST is capable of detecting such clues by extrapolating features from past sequences, which also implicitly contains the camera motion, without training a pose estimation network conditioned to specific future frames, like in \cite{nag2022far}, that clearly limits the application to forecast depths only at corresponding future time steps. We report a slight drop in terms of RMSE at mid-term compared to \cite{hu2020probabilistic} and \cite{nag2022far}, however we still achieve concrete improvements in terms of RMSE-Log, by reducing the error of 22\%. This indicates that the relative depth consistency is much better preserved by our approach than by the competitors. 
%At mid-term, FLODCAST exceeds all supervised methods, with drops only in terms of RMSE error. We get huge improvements in terms of both the AbsRel and SqRel errors with very accurate pixels according to the inlier percentages. We also see that FLODCAST performs well in comparison with DeFNet: we mainly found higher RMSE error, while keeping lower the AbsRel (-36\%), the SqRel (-5\%) and the logarithmic RMSE, with high accuracy as seen for inliers results.

%In general, we designed a very competitive approach able to reduce the absolute error while maintaining as higher as possible the accuracy of anticipated depth maps, even after many time steps, without training the network for some specific time instants. This allows us to address the depth anticipation problem at any $t+k$ time instant in the future.

%In general, we designed a simple and effective approach able to forecast depth maps, by using optical flows and depth maps of past video frames, at the highest accuracy and the lowest error up to $t+10$ frame ahead, as show in Tab \ref{table:depthresults}.
Using its recurrent nature, FLODCAST is capable to generate a sequence of depth maps in the future without temporal sub-sampling, i.e. by producing all the intermediate forecasting steps (not only the last one, as done in \cite{nag2022far}).
In dynamic scenarios, like an urban setting, this is particularly useful, since objects can appear and be occluded several times from one frame to another. Such behavior might not emerge from subsampled predictions.

%In Fig. \ref{fig:depthtestset}, we report the AbsRel error of FLODCAST both at short-term and mid-term predictions grouped by depth values. We observed that FLODCAST makes most of the errors at the first meters, mainly due to few depth data annotated in the dataset, as it can be clearly seen by looking at the occurrences histogram shown in Fig. \ref{fig:occhisttest}.
%
%\begin{figure}[t]
%\includegraphics[width=\linewidth]{images/absrel_hist_test_set.png}
%\caption{AbsRel error distribution on Cityscapes test set both at short-term (blue bars) and mid-term (orange bars). Pixel occurrences histogram is depicted in Fig. \ref{fig:occhisttest}.}\label{fig:depthtestset}
%\end{figure}
%

%
%
\begin{figure}[t]
\resizebox{0.93\columnwidth}{!}{\begin{minipage}{\textwidth}
\begin{tabular}{ccc}
\toprule
\multicolumn{3}{c}{Short-term $k=5$} \\ \toprule
RGB & Depth GT & Our Predictions \\
\includegraphics[width=.33\linewidth]{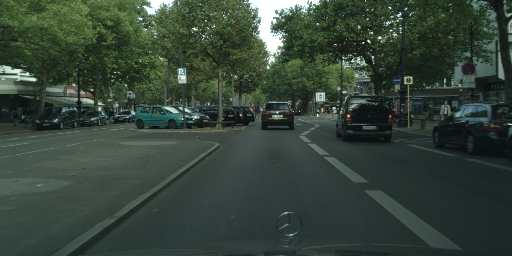} & \includegraphics[width=.33\linewidth]{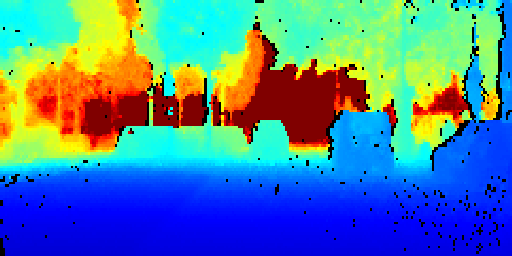} & \includegraphics[width=.33\linewidth]{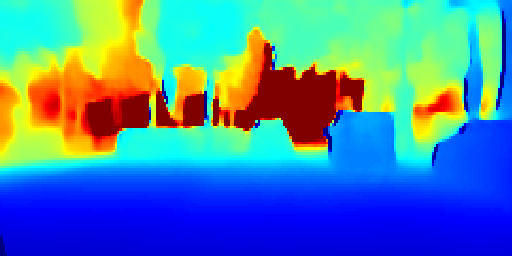} \\ 
\includegraphics[width=.33\linewidth]{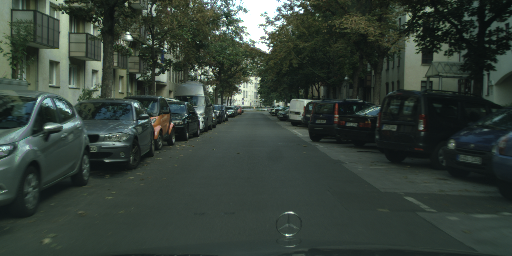} & \includegraphics[width=.33\linewidth]{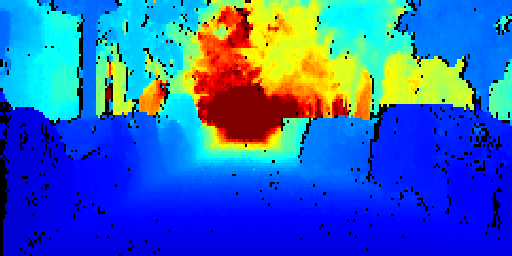} & \includegraphics[width=.33\linewidth]{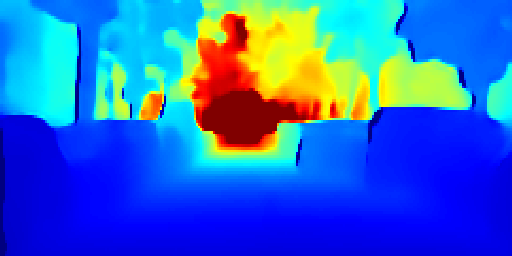} \\
\includegraphics[width=.33\linewidth]{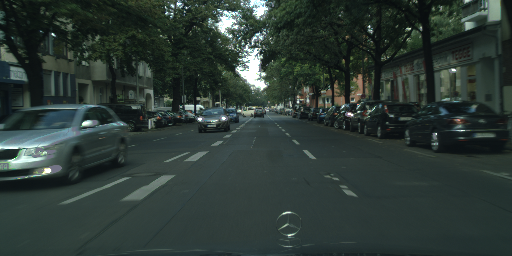}  & \includegraphics[width=.33\linewidth]{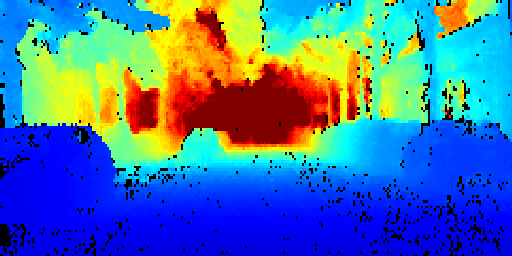} & \includegraphics[width=.33\linewidth]{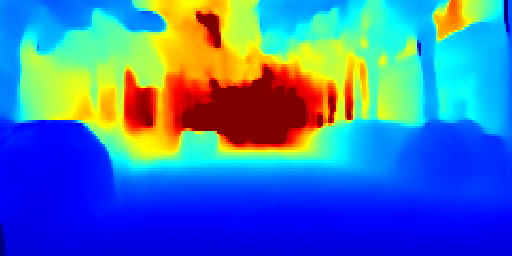} \\ \bottomrule
\end{tabular}
\caption{Visualization results of future predictions on Cityscapes test set at short-term. Black pixels in the ground truth (second column) are invalid measurements.}\label{fig:shortdepthresults}
\end{minipage}}
\end{figure}

\begin{figure}[t]
\resizebox{0.93\columnwidth}{!}{\begin{minipage}{\textwidth}
\begin{tabular}{ccc}
\toprule
\multicolumn{3}{c}{Mid-term $k=10$} \\ \toprule
RGB & Depth GT & Our Predictions \\
\includegraphics[width=.33\linewidth]{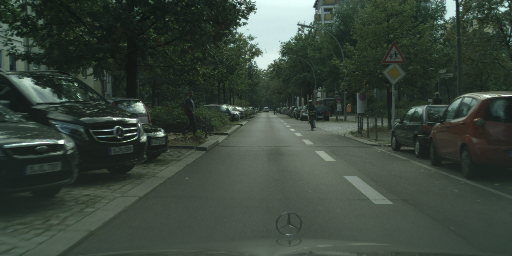} & \includegraphics[width=.33\linewidth]{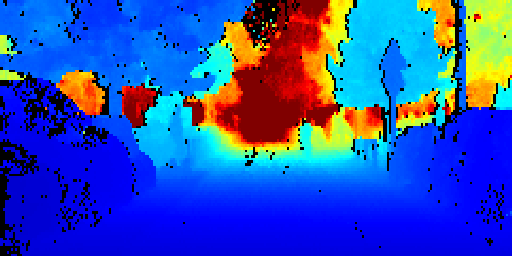} & \includegraphics[width=.33\linewidth]{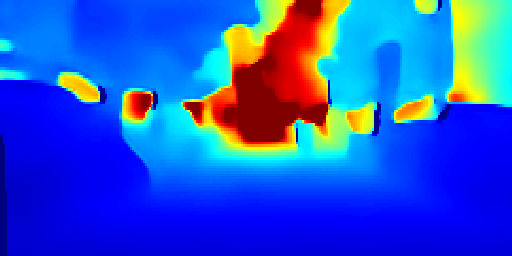} \\ 
\includegraphics[width=.33\linewidth]{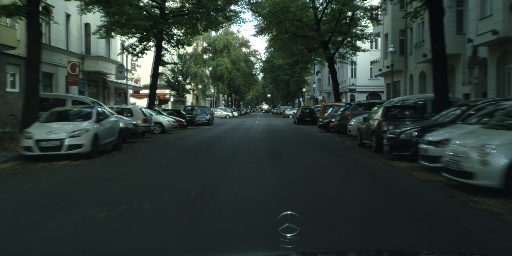} & \includegraphics[width=.33\linewidth]{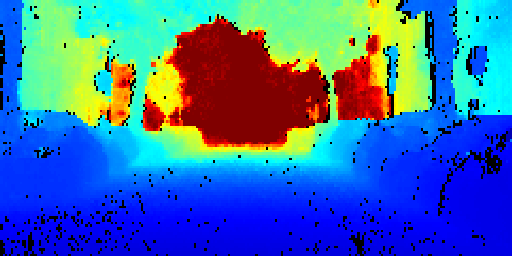} & \includegraphics[width=.33\linewidth]{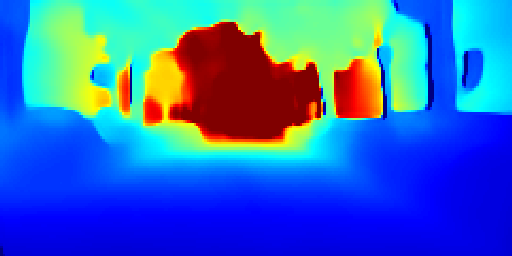} \\
\includegraphics[width=.33\linewidth]{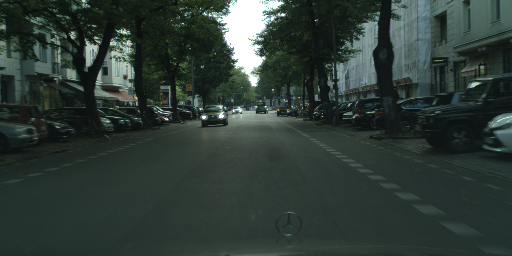}  & \includegraphics[width=.33\linewidth]{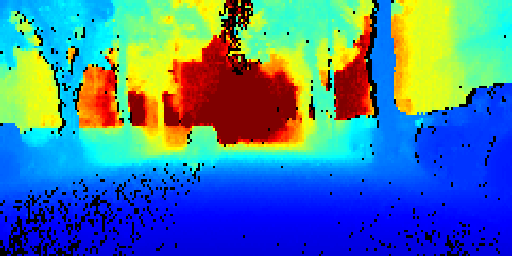} & \includegraphics[width=.33\linewidth]{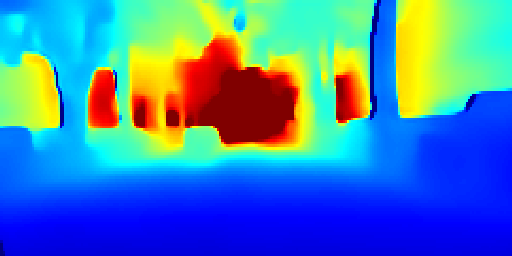} \\ \bottomrule
\end{tabular}
\caption{Visualization results of future predictions on Cityscapes test set at mid-term. Black pixels in the ground truth (second column) are invalid measurements.}\label{fig:middepthresults}
\end{minipage}}
\end{figure}

%Although this lack of annotated data, the overall depth maps result very accurate. 
Some qualitative results are shown in Fig. \ref{fig:shortdepthresults} and \ref{fig:middepthresults}, respectively for short-term and mid-term predictions.
FLODCAST learns to locate the region containing the vanishing point by assigning higher depth values.
%that lies on the center of the image, since Cityscapes sequences are recorded from an ego-vehicle perspective, therefore it correctly assigns on pixel area around to the vanishing point higher depth values (i.e. colored "red" indicating "very far away"). %On the other hand, top and bottom of the long sides of the frame are usually annotated as close. 
Moreover, we observed that missing depth map values coming from zeroed values in the ground truth frames are mostly predicted correctly. This underlines that FLODCAST is able to anticipate depth maps up to mid-range predictions while being highly accurate, even though some parts of the scene may not have been labeled, due to bad measurements or missing data.%(more quantitative details are in AbsRel errors histogram, see Fig. \ref{fig:depthtestset})

%

%-------------<
\subsection{Future Flow Estimation}\label{sec:futflowest}

We evaluate optical flow forecasting capabilities on Cityscapes, by following the protocol of \cite{jin2017predicting}. Therefore, we calculate the average end-point error EPE, according to Eq. \ref{eq:epeflow}, for the $t+10$ frame (i.e. 0.59 sec ahead), namely corresponding at the 20th frame for each val sequence. We carry out experiments at the resolution $256 \times 512$, by doubling the resolution, and we compare our approach with existing works, FAN~\cite{jin2017predicting} and OFNet~\cite{ciamarra2022forecasting}, and some baselines from \cite{jin2017predicting}, namely (i) warping the flow field using the optical flow in each time step (namely \textit{Warp Last}) and (ii) simply copying the one last (namely \textit{Copy Last}).

%We evaluate our results in terms of flow forecasting capabilities, by following previous work~\cite{ciamarra2022forecasting}. Therefore, we measure our optical flow predictions generated autoregressively for each time step, by computing the mean squared error for the $u$ and $v$ components and averaging them, according to Eq. \ref{eq:mseflow}. 

%We compute flow predictions at mid-term, i.e. up to $t+9$ (after 0.5s), by feeding to FLODCAST $\{OF_{8}, OF_{9}, OF_{10}\}$ and $\{D_{8}, D_{9}, D_{10}\}$ and we measure the errors at $128 \times 256$ for every steps in an autoregressive way, at the 20th frame of each video sequence in the validation set, i.e. following the same evaluation strategy of~\cite{luc2017predicting, luc2018predicting, ciamarra2022forecasting}. We first compare our approach with OFNet, since we want to assess the flow performance for every intermediate steps. 

Since our work is capable to provide optical flows for multiple future scenarios, we also assess our performance for every intermediate frames up to $t+10$, by following the evaluation protocol in \cite{ciamarra2022forecasting}. Thus, we measure the quality of our predictions generated autoregressively for each time step, by computing the mean squared error for $u$ and $v$ components and averaging them, according to Eq. \ref{eq:mseflow}. We report our quantitative results in Tab.~\ref{tab:flowresults}.

\begin{table}[t]
\caption{Qualitative results for flow forecasting on Cityscapes val set. In bold the lowest error. We denote with the symbol ``$-$'' if the corresponding result is not available or reproducible.}\label{tab:flowresults}
\resizebox{\linewidth}{!}{
\begin{tabular}{l|cccccccccc|c}
\toprule
\multirow{2}{*}{Method} & \multicolumn{10}{c|}{MSE $\downarrow$} & EPE $\downarrow$ \\ \cline{2-12}
 & t+1 & t+2 & t+3 & t+4 & t+5 & t+6 & t+7 & t+8 & t+9 & t+10 & t+10 \\
\toprule
Copy Last~\cite{jin2017predicting} & $-$ & $-$ & $-$ & $-$ & $-$ & $-$ & $-$ & $-$ & $-$ & $-$ & 9.40 \\
Warp Last~\cite{jin2017predicting} & $-$ & $-$ & $-$ & $-$ & $-$ & $-$ & $-$ & $-$ & $-$ & $-$ & 9.40 \\
FAN~\cite{jin2017predicting} & $-$ & $-$ & $-$ & $-$ & $-$ & $-$ & $-$ & $-$ & $-$ & $-$ & 6.31 \\

% OFNet~\cite{ciamarra2022forecasting} & 0.72 & 1.08 & 1.20 & 1.56 & 1.68 & 1.98 & 2.16 & 2.56 & 2.38 & 
%  & 2.08 \\
% FLODCAST  w/o depth & \textbf{0.66} & \textbf{0.92} & \underline{1.02} & \underline{1.38} & \underline{1.32} & \underline{1.56} & \underline{1.76} & \underline{2.00} & 1.70 &  & 1.48 \\
% FLODCAST (Ours)  & \underline{0.70} & \underline{0.94} & \textbf{1.00} & \textbf{1.30} & \textbf{1.30} & \textbf{1.50} & \textbf{1.64} & \textbf{1.98} & \textbf{1.62} &  & \textbf{1.38} \\

OFNet~\cite{ciamarra2022forecasting} & \textbf{0.96} & 0.94 & 1.30 & 1.40 & 1.78 & 1.88 & 2.16 & 2.38 & 2.88 & 2.66 & 2.08 \\
FLODCAST  w/o depth & 0.98 & \textbf{0.80} & 1.11 & 1.20 & 1.38 & 1.48 & 1.72 & 1.78 & 2.18 & 1.92 & 1.48 \\
FLODCAST (Ours)  & 1.06 & 0.84 & \textbf{1.10} & \textbf{1.12} & \textbf{1.34} & \textbf{1.44} & \textbf{1.62} & \textbf{1.68} & \textbf{2.12} & \textbf{1.74} & \textbf{1.38}  \\
\bottomrule
\end{tabular}}
\end{table}

%
%\begin{table}[t]
%\centering
%\caption{Quantitative results for flow forecasting in Cityscapes val set, in terms of mean squared error computed between the two flow channel them averaging them. We show that SceMCA drastically reduced prediction errors through the time. We indicate in bold the lowest error at every time step.}\label{tab:flowresults}
%\resizebox{\linewidth}{!}{%\begin{minipage}{\linewidth}
%\begin{tabular}{l|ccccccccc}
%\toprule
%\multirow[t]{2}{*}{Method} & \multicolumn{9}{c}{MSE $\downarrow$} & \cline{2-10} %\multicolumn{1}{l|}{Method} 
%& ~t+1~ & ~t+2~ & ~t+3~ & ~t+4~ & ~t+5~ & ~t+6~ & ~t+7~ & ~t+8~ & ~t+9~\\ \midrule
%\multicolumn{1}{l|}{OFNet\cite{ciamarra2022forecasting}} & \textbf{0.36} & \textbf{0.54} & 0.60 & 0.78 & 0.84 & 0.99 & 1.08 & 1.28 & 1.19\\
%\hline
%%\multicolumn{1}{l|}{SceMCA w/o depth} & 0.51 & 0.54 & 0.54 & 0.78 & 0.74 & 0.79 & 0.92 & 1.09 & 0.88\\
%\multicolumn{1}{l|}{SceMCA (Ours)} & 0.49 & 0.56 & \textbf{0.53} & \textbf{0.74} & \textbf{0.69} & \textbf{0.75} & \textbf{0.88} & \textbf{1.02} & \textbf{0.80}\\
%\bottomrule
%\end{tabular}
%%\end{minipage}
%}
%\end{table}
%
We mainly found that the FLODCAST error drastically decreases over time. This brings us some considerations. Fist of all, FLODCAST combines different modalities, also exploiting spatio-temporal information, and that comes to be crucial to reduce the accumulation error through time. Because optical flow and depth maps are complementary each other, the model can better identify specific patterns, e.g. discriminating object motions at different resolutions in advance (see Fig. \ref{fig:flowqualitativeres}). This also allows to directly generate multiple future optical flows at a time with a shorter input sequence (i.e. $T=3$ for FLODCAST while $T=6$ for OFNet). Moreover, we found a substantial diminishing of the MSE up to 33\% at $t+10$ and that also supports our observations. 
Considering that OFNet has more supervision during training, i.e. it forecasts an output sequence shifted by one step ahead with respect to its input, this is the reason we believe performances are sometimes better at the beginning steps but then the error increases compared to FLODCAST.

%FLODCAST exploits flow and depth features, which are combined to enrich pixel-level information through spatio-temporal contexts, and that comes to be crucial to reduce the accumulation error through the time. This also allows to directly generates multiple future optical flows at a time with a shorter input sequences (i.e. $T=3$ for FLODCAST, $T=6$ for OFNet). Moreover, we found a substantial diminishing of the MSE up to +30\% at $t+10$. Considering that OFNet has more supervision during the training, i.e. it forecasts an output sequence shifted by one step ahead with respect to its input, this is the reason we believe performances are sometimes better at the beginning steps but then the error diverges much more than FLODCAST.

%We also calculate the average end-point error EPE\cite{baker2011database}, which is defined as $\text{EPE}:=\frac{1}{N}\sqrt{(u-u_{GT})^2 + (v-v_{GT})^2}$ over $N$ pixels images,  

In absence of intermediate results of MSE for other methods (i.e. FAN, for which no source code and models are available, as denoted in Tab. \ref{tab:flowresults}), we compare the overall performance by evaluating the EPE error at $t+10$, also against the Flow Anticipating Network (FAN) proposed in \cite{jin2017predicting}, that generates future flows in a recursive way, by using the finetuned version of their model, which is learned to predict the flow for the single future time step given the preceding frames and the corresponding segmentation images.
%We also calculate the average end-point error EPE (see Eq. \ref{eq:epeflow}) so to compare our approach with the Flow Anticipating Network (FAN) proposed in \cite{jin2017predicting}, that generates future flows in a recursive way, by using the finetuned version of their model, which is learned to predict the flow for the single future time step given the preceding frames and the corresponding segmentation images. To compare our results with other works, we doubled the resolution of our predictions. Hence, we report in Tab. \ref{tab:flowresults} the flow performance at $t+10$, together with our FLODCAST, OFNet and some baselines\cite{jin2017predicting}, which includes (i) warping the flow field using the optical flow in each time step (namely \textit{Warp Last}) and (ii) simply copying the one last (namely \textit{Copy Last}).
%that generates the future flow in a recursive way. Hence, we report in Tab. \ref{tab:epeflow} the flow performance at $t+10$.
%
%\begin{table}[h]
%\centering
%\parbox{.4\linewidth}{
%\caption{Average end-point error (EPE) on Cityscapes val set at $t+10$.}\label{tab:epeflow}
%\resizebox{0.9\linewidth}{!}{
%\begin{tabular}{l|c}
%\toprule
%Method & EPE $\downarrow$\\
%\toprule
%Copy Last Input \cite{jin2017predicting} & 9.40\\
%Warp Last Input \cite{jin2017predicting} & 9.40\\
%\hline
%FAN \cite{jin2017predicting} & 6.31\\
%OFNet \cite{ciamarra2022forecasting} & 2.08\\
%\hline
%SceMCA w/o depth & 1.60\\
%SceMCA (Ours) & \textbf{1.48}\\
%\bottomrule
%\end{tabular}}}
%}}
%\end{table}
%

We found remarkable improvements even at $t+10$, by reducing the EPE with respect to FAN and OFNet as well. This highlights our choice that using optical flow with depth maps is better for determining future estimates than with the semantic segmentations employed in FAN. Restricting to observing past optical flows to generate a future one, as done in OFNet, does not allow forecasting models to make reliable long-range predictions autoregressively. Further improvements are obtained when multiple frames are predicted at a time, as FLODCAST does.
Then, we demonstrate that FLODCAST is more accurate in predicting unobserved motions far into the future, without requiring semantic data, that is typically harder to get labeled with respect to depth maps, which are directly obtained by using commercial devices like LiDARs or stereo rigs. We also observe that excluding the depth map from FLODCAST, flow performance is reduced, since EPE increases by 6.8\%. Despite the hard task of anticipating flow motion without seeing future frames, FLODCAST exceeds all the previous works, and it is more robust when depth is stacked into the input data.
%Further observations will be provided with the ablation study (see paragraph \ref{sec:flowablationstudy}).

\subsection{Ablation Study}\label{sec:ablationstudy}

%We conduct an ablation study to explore if either optical flow or depth as inputs can weaken the predictions, by investigating benefits valuable for future predictions.
In order to understand how significant the flow and depth as data sources are for anticipating the future, we exclude one of the two inputs at a time and we evaluate the performance compared with FLODCAST, which instead leverages both data sources.

% Depth
\paragraph{Depth Analysis}\label{sec:depthablationstudy}

To demonstrate the importance of incorporating flow features for depth forecasting, we exclude optical flow from the input and we train FLODCAST using the $\mathcal{L}_{\text{depth}}$ loss (see Eq. \ref{eq:depthlossfunction}) to estimate future depth maps.

\begin{figure}[t]
\centering
\includegraphics[width=\linewidth]{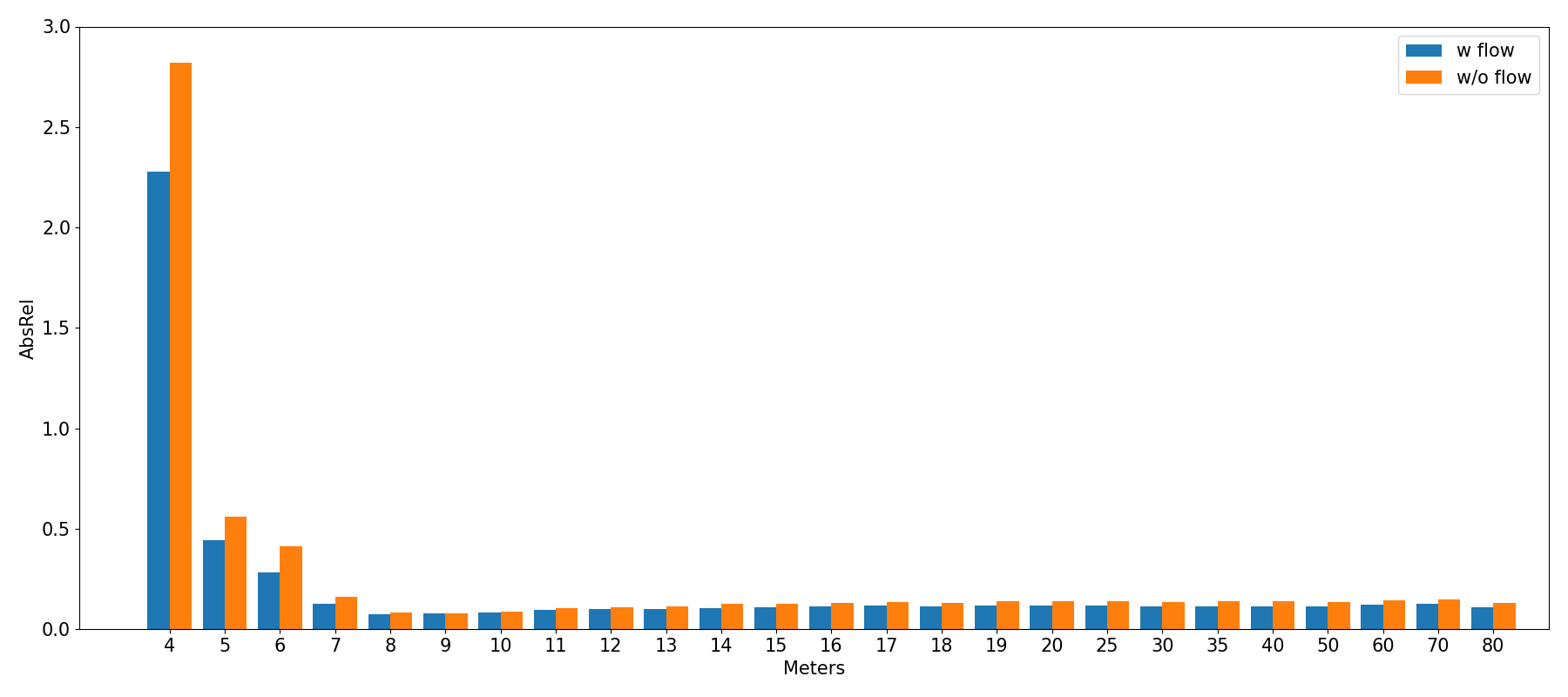}
\caption{Ablation study on depth forecasting in Cityscapes test set. We report the AbsRel error at $t+10$ per distance (in meters), both when the input data is composed of optical flows and depth maps (blue) or only depth (orange). Note that depth values below 3 meters are not present in the test set.}\label{fig:scatterplotdepth}
\end{figure}

From Tab. \ref{table:depthresults} we observe that generating future depth maps through the past ones without leveraging optical flow as source data, i.e. FLODCAST w/o flow, worsens the predictions under all of the metrics. This points out the relevance of combining features extracted from past scenes, in terms of 2D motion and depth. Nonetheless, predicting only future depth maps using our approach, even discarding the optical flow information, gets improvements compared to prior works such as \cite{hu2020probabilistic, nag2022far}. At short-term $t+5$ FLODCAST w/o flow is the second best result overall, by reducing the errors by a large margin (e.g. AbsRel and SqRel respectively -53\% and -27\% from Hu et al, and -52\% and -16\% from DeFNet) with also higher percentage of inliers. At mid-term $t+10$ we reported drops of performance of FLODCAST w/o flow still limiting the AbsRel error and getting higher accuracy of inlier pixels. Overall, removing optical flow from the input data, FLODCAST still works better than all the existing works on forecasting unseen scenarios but then the lack of the information affects the performance for farther frames.
%demonstrating that the proposed architecture is suitable to anticipate depth maps far into the future. 
In addition, we compute the AbsRel error distribution of FLODCAST, when depth maps are predicted through only optical flows (orange bars) or employing our multimodal approach (blue bars) and we plot a histogram at $t+10$ as function of the distance (Fig. \ref{fig:scatterplotdepth}).
%
%\begin{figure}[t]
%\resizebox{\columnwidth}{!}{
%\begin{tabular}{cc}
%\includegraphics[width=.5\linewidth]{images/ablation/depth/t+10/scatter_1000_absrel.png} & \includegraphics[width=.5\linewidth]{images/ablation/depth/t+10/scatter_1000_absrel_wo_flow.png} \\
%\noalign{\smallskip} (a) SceMCA & (b) SceMCA w/o flow \\
%\end{tabular}}
%\caption{Ablation study on depth forecasting in Cityscapes test set. We report pixel-wise AbsRel errors over the distance (in meters) of depth maps predicted by SceMCA, when the model takes as input both the optical flows and depth maps (a) or only depth maps (b).}\label{fig:scatterplotdepth}
%\end{figure}
%
%

We found notable improvements within 10 meters when optical flow is part of the input. This is crucial in terms of safety since objects moving around a self-driving agent can be better defined according to their predicted distances. Indeed, from an ego-vehicle perspective, parts of the scene close to the observer are more likely to change over time. Considering that we are forecasting the depth for the whole image, just a few regions move considerably, corresponding to dynamic objects. The rest of the scene, typically the background, like buildings or vegetation, exhibits instead a static behavior and does not change much depth-wise even in presence of ego-motion. Therefore, the depth estimated for those far away pixels contains little error and, consequently, the tails of the two plots tend to be quite similar. Considering that the histogram represents depth errors 10 frames after from the last observed one, our FLODCAST is robust also for long distance when optical flow is part of the input. This also motivates our design choices of sticking data in a multimodal and multitasking approach.
 
\begin{figure}[t]
\includegraphics[width=.286\columnwidth]{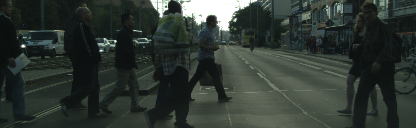} ~~~ \includegraphics[width=.286\columnwidth]{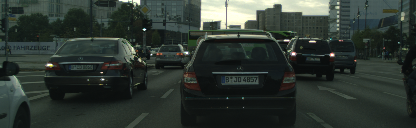} ~~~
\includegraphics[width=.286\columnwidth]{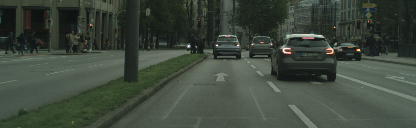} \\
\includegraphics[width=.286\columnwidth]{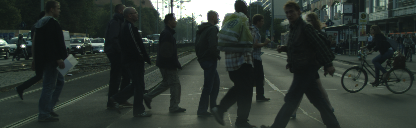} ~~~ \includegraphics[width=.286\columnwidth]{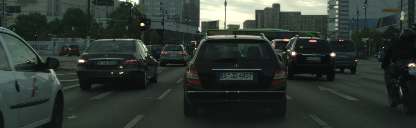} ~~~ \includegraphics[width=.286\columnwidth]{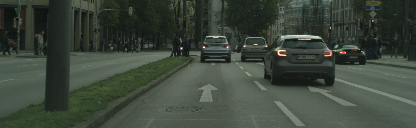} \\
\includegraphics[width=.286\columnwidth]{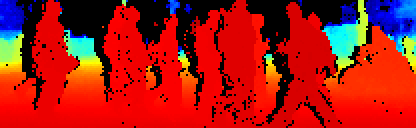} ~~~  \includegraphics[width=.286\columnwidth]{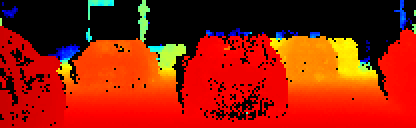} ~~~ 
\includegraphics[width=.286\columnwidth]{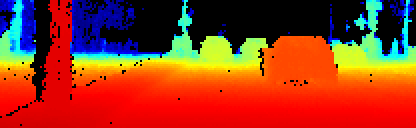} \\
\includegraphics[width=.286\columnwidth]{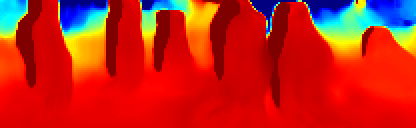} ~~~  \includegraphics[width=.286\columnwidth]{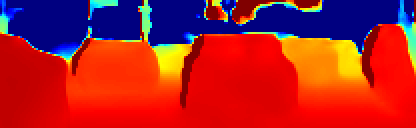} ~~~ 
\includegraphics[width=.286\columnwidth]{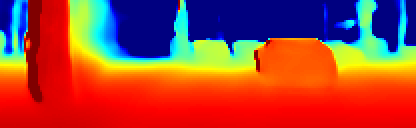} \\
\includegraphics[width=.286\columnwidth]{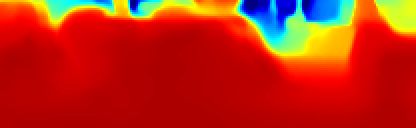} ~~~ \includegraphics[width=.286\columnwidth]{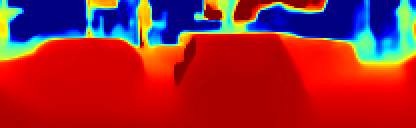} ~~~ 
\includegraphics[width=.286\columnwidth]{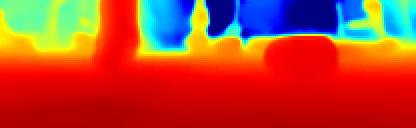} \\
\includegraphics[width=.329\linewidth]{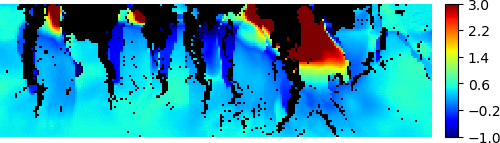} \includegraphics[width=.329\linewidth]{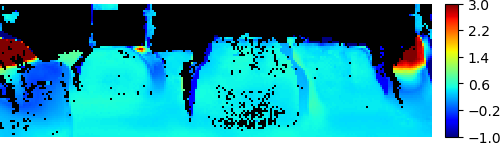}
\includegraphics[width=.329\linewidth]{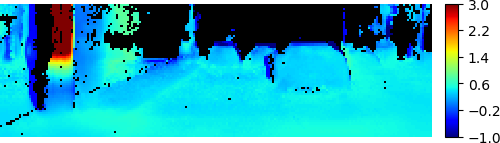}
\caption{Qualitative results of predicted depth maps of FLODCAST trained with or without optical flows (4th and 5th row respectively). The first two rows are the last observed frame $I_t$ and the future one, $I_{t+10}$. The third row contains ground truth depth maps for the three samples. Pixel-wise AbsRel errors between FLODCAST w/o flow and our FLODCAST are depicted as heatmap plots in the 6th row for 3 different sequences in the Cityscapes test set.}\label{fig:depthqualitativeres}
\end{figure}

We further provide some qualitative results in Fig. \ref{fig:depthqualitativeres}, so to underline how the contribution coming from the flow features is significant in generating very accurate depth maps, especially on moving objects, like pedestrians and vehicles. 
It is noteworthy that 2D motion displacements in the scene help to correctly predict depth values on different moving objects close to each other, e.g. pedestrians crossing the street, whose estimated depths collapse in a unique blob when optical flow is not taken into account. The same happens for cars at different distances from the camera, where their predicted depths look lumped together. That suggests that the model without flow features is less capable of distinguishing single instances.%, either obstacles that potentially might come, e.g. the pole is not completely defined in the third example (see the right side of the image).

% Flow
\paragraph{Flow Analysis}\label{sec:flowablationstudy}

\begin{figure}[t]
\centering
\includegraphics[width=\linewidth]{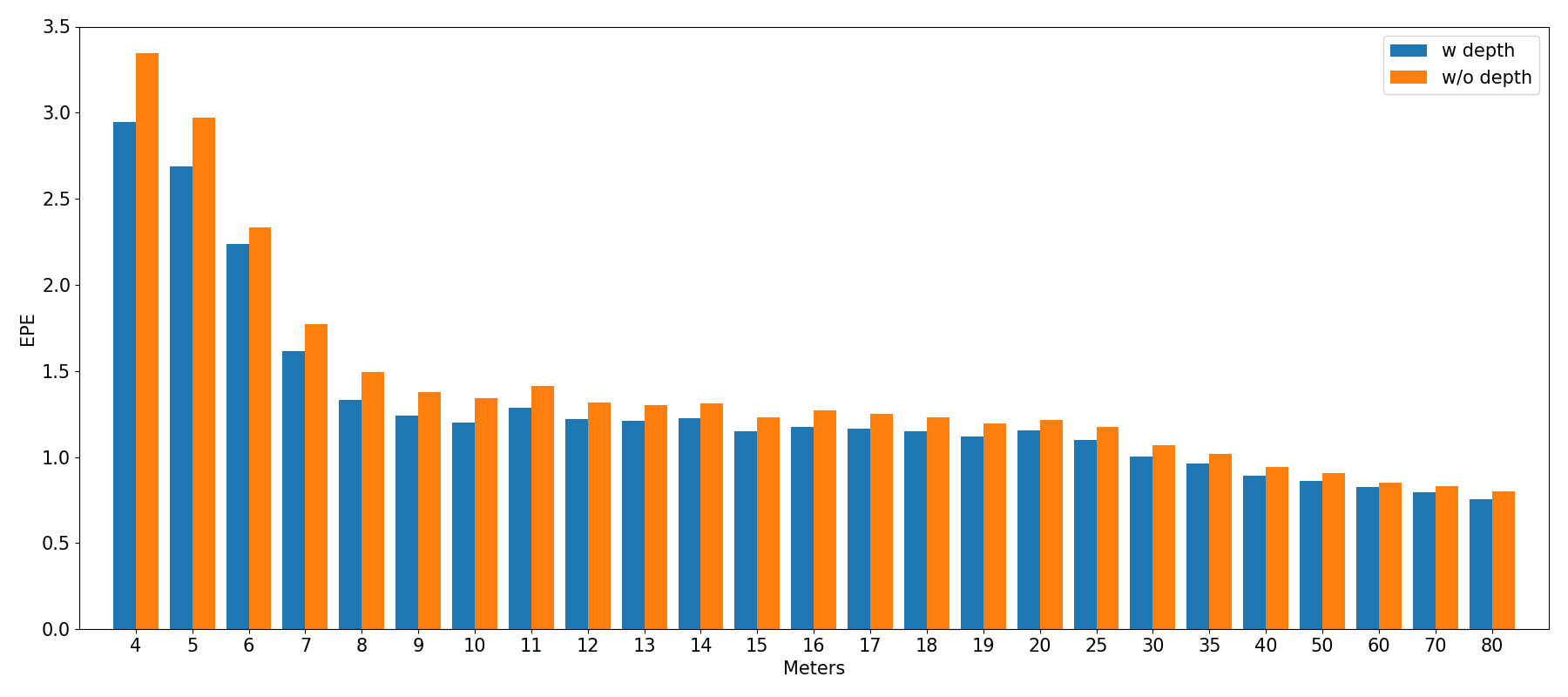}
\caption{Ablation study on flow forecasting in the Cityscapes test set. We report the EPE error at $t+10$ according to the distance (meters) of optical flows predicted by FLODCAST, in case of the input data being both optical flows and depth maps (blue) or only optical flows (orange). Note that depth values below 3 meters are not present in the test set.}\label{fig:scatterplotflow}
\end{figure}

We discard depth maps from the input data and we train the network to predict future optical flows, i.e. by exploiting past flow features, while keeping the same $\mathcal{L}_{\text{flow}}$ loss (see Eq. \ref{eq:flowlossfunction}). We measure the optical flow predictions generated autoregressively for each time step, by computing the mean squared error on both the two flow channels and averaging them (Eq. \ref{eq:mseflow}).
From the flow forecasting results reported in Tab. \ref{tab:flowresults}, we observe that features extracted from both the optical flows and depth maps contribute to reduce the MSE errors on predicted flows, resulting in overall improvements after the first steps up to at $t+10$, i.e. +33\% over OFNet and +9\% over FLODCAST w/o depth, which is significant considering the high uncertainty for farther future scenarios. Compared with OFNet, FLODCAST w/o depth has the FlowHead module (as depicted in Fig. \ref{fig:scemca}), in which specialized weights of convolutional layers are end-to-end trained in order to directly generate multiple optical flows at a time. Despite the notable reduction of the error through time, FLODCAST overcomes its performance when depth maps are included in the source data, which points out the importance of our multimodal approach. %Besides, our forecasting network, which is designed to generate multiple predictions simultaneously, is prone to generate more accurate results than other existing approaches. 
Looking at the last prediction, i.e. at $t+10$, FLODCAST w/o depth still exceeds other approaches, but reports an increase of the EPE error by +7\% with respect to our multimodal approach. This fact suggests that recurrent architectures can achieve good results for forecasting tasks and they can improve if they are multimodal.
In addition, we study the EPE error distribution according to distance.
%using a bar histogram and we will discuss improvements of FLODCAST with or without depth map. 
To do that, we collect all the predicted flows upsampled to $256 \times 512$ at $t+10$ on the test set, and we compute the error (see Eq. \ref{eq:epeflow}) for all the pixels falling into the corresponding distance-based bins and we represent their averages in Fig. \ref{fig:scatterplotflow}. Here, orange bars are errors reported by only using optical flow in input, while the blue ones incorporate also depth maps, i.e. our proposed  FLODCAST model.
%In addition, we study the EPE error distribution according to the distance. To do that, we collect all the predicted flows upsampled to $256 \times 512$ at $t+10$ on the test set, and we compute the error (see Eq. \ref{eq:epeflow}), which is equivalent to the euclidean distance between the two flow vectors (the prediction and its groundtuth) for all the pixels falling into a certain distance and we represent its average value as a single bar of a histogram. We do that, using FLODCAST in case we include or not depth map together with optical flow as data source, in order to find out how are distributed the improvements in relation to the distance from the camera. Our results are depicted in Fig. \ref{fig:scatterplotflow}, where orange bars are errors reported using FLODCAST with optical flow only, while the blue ones incorporate also depth map.
%
%\begin{figure}[t]
%\resizebox{\columnwidth}{!}{
%\begin{tabular}{cc}
%\includegraphics[width=0.5\linewidth]{images/ablation/flow/t+10/scatter_1000_mse.png} & \includegraphics[width=0.5\linewidth]{images/ablation/flow/t+10/scatter_1000_mse_wo_depth.png} \\
%(a) SceMCA & (b) SceMCA w/o depth \\
%\end{tabular}}
%\caption{Ablation study on flow forecasting in Cityscapes test set. We report the pixel-wise MSE error according to the distance (in meters) of optical flows predicted by SceMCA, in case of the input data is both the optical flows and depth maps (a) or only optical flows (b).}\label{fig:scatterplotflow}
%\end{figure}
%
%

As can be seen in Fig. \ref{fig:scatterplotflow}, the overall trend of EPE is to decrease as the depth increases. This is due to the fact that, parts of the scene far enough from the camera typically produce similar small motion, like objects moving at the background or static parts that are mainly affected by the camera motion, thus the predicted optical flows for such pixels are likely to be more accurate. Instead, pixels closer to the camera tend to have a more pronounced motion and that affects the predictions, especially of farther frames. We observe that EPE errors of FLODCAST are always lower when depth maps are provided as input (blue bars) than only using optical flow as unique data source (orange bars). In particular, we gain more within 15 meters, which is the most relevant part of the scene concerning the safety and the drive planning of autonomous agents in very dynamic scenarios like the urban one. FLODCAST with depth maps has the potential to better disambiguate motions of pixels close to the observer than the far ones and vice versa.
%, and that is the more relevant the more the future becomes uncertain and harder.
%This ablation study have shown that using depth map with optical flow is always preferable to better estimate optical flows in advance.

\begin{figure}[t]
\includegraphics[width=.286\linewidth]{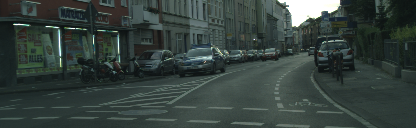} ~~~ \includegraphics[width=.286\linewidth]{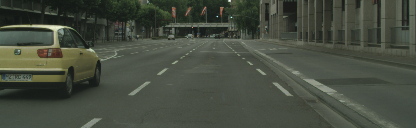} ~~~
\includegraphics[width=.286\linewidth]{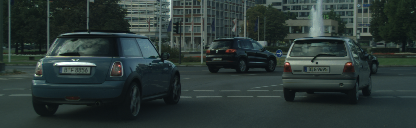} \\
\includegraphics[width=.286\linewidth]{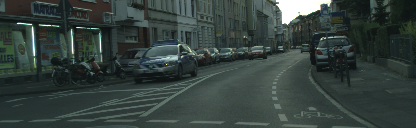} ~~~  \includegraphics[width=.286\linewidth]{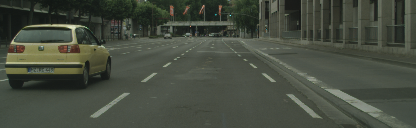} ~~~ 
\includegraphics[width=.286\linewidth]{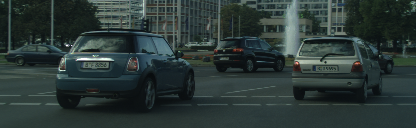} \\
\includegraphics[width=.286\linewidth]{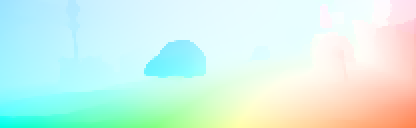} ~~~ \includegraphics[width=.286\linewidth]{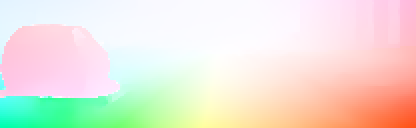} ~~~ 
\includegraphics[width=.286\linewidth]{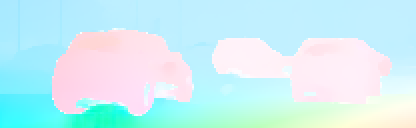} \\
\includegraphics[width=.286\linewidth]{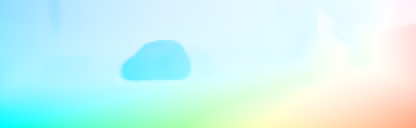} ~~~ \includegraphics[width=.286\linewidth]{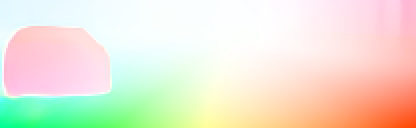} ~~~
\includegraphics[width=.286\linewidth]{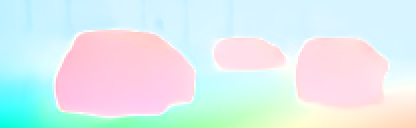} \\
\includegraphics[width=.286\linewidth]{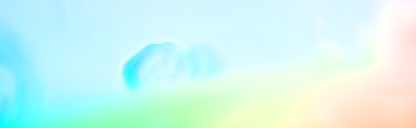} ~~~ \includegraphics[width=.286\linewidth]{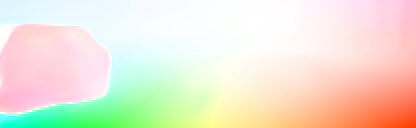} ~~~
\includegraphics[width=.286\linewidth]{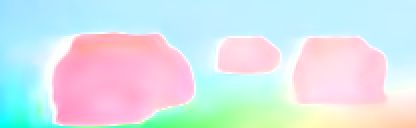} \\
\includegraphics[width=.329\linewidth]{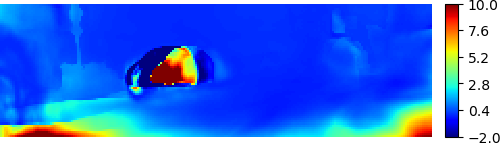}  \includegraphics[width=.329\linewidth]{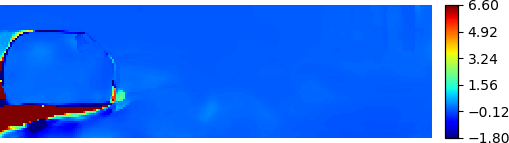}
\includegraphics[width=.329\linewidth]{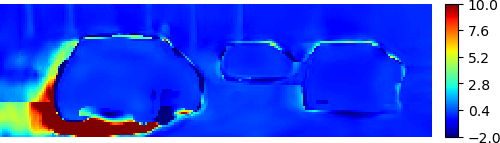}
\caption{Flow forecasting qualitative results on the Cityscapes test set. We use FLODCAST trained with or without depth maps (4th and 5th row respectively). The first two rows depict the last observed frame $I_t$ and the future one, $I_{t+5}$. The third row shows ground truth flows. In the 6th row we depict the difference of MSE errors wrt the ground truth between the predictions of FLODCAST using only past flows and both past flows and depths.}\label{fig:flowqualitativeres}
\end{figure}

%TODO: da riadattare le considerazioni con questo nuovo plot.
%The MSE errors depicted in blue are more spread along the distances, even few meters far from the camera because the depth map is discarded. This is also due to the fact that displacements of objects moving that are even close to the camera are more accurately estimated when the distance is known. 
Hence, flow forecasting results are more precise as long as the depth map is included in the input data. Based on this consideration, we reported in Fig. \ref{fig:flowqualitativeres} some qualitative results on the Cityscapes test set, where we illustrate the ground truth optical flow in comparison with the optical flows obtained from FLODCAST, both exploiting or not the depth map as an additional input source.
%i.e. respectively 3rd, 4th and 5th rows.
Finally, we show the heatmaps in the last row of Fig. \ref{fig:flowqualitativeres} of the MSE errors with respect to the ground truth as differences between the predictions generated by FLODCAST without depth map and by FLODCAST using both data sources. Specifically, we report enhancements mostly on moving objects, whose shapes are more correctly defined, as shown in the red parts of the cars and the light blue around their shapes.

\subsection{Performance details}\label{sec:performance}

To take into account the forecasting problem in terms of anticipation, predictions have to be provided early. We therefore analyse the performance of FLODCAST at inference time. We test our model using a single NVIDIA RTX 2080.
%We are considering to anticipate not only a single piece of information, i.e. the depth as in \cite{nag2022far}, but also the optical flow as in \cite{jin2017predicting, qi20193d, hu2020probabilistic}, %so to encapsulate the scene motion in a single architecture.
%so to provide meaningful semantic-level details of the scene that has yet to come.
At runtime, FLODCAST requires 8.8GB of GPU memory and it is able to forecast sequences of $K=3$ consecutive depth maps and optical flows in 40ms (25FPS). Our predictions are estimated for multiple frames ahead simultaneously, which is more efficient than making predictions for a single one, as done in \cite{jin2017predicting, qi20193d, hu2020probabilistic}.
%As far as we know, forecasting optical flows and depth maps for multiple future frames at a time in a single architecture was not explored in other existing works.

%\appendix
%\section{Video Prediction}\label{sec:videoprediction}
\section{Segmentation Forecasting}\label{sec:videoprediction}

We now show how FLODCAST can be employed to address downstream tasks such as forecasting segmentation masks.
In fact, flow-based forecasting methods have demonstrated that warping past features onto future frames allows producing competitive semantic segmentations~\cite{terwilliger2019recurrent,  saric2020warp, ciamarra2022forecasting}. Since FLODCAST predicts dense optical flows in the future, we use the recent lightweight framework introduced in \cite{ciamarra2022forecasting}, to explore possible improvements on the segmentation forecasting problem as a downstream task through our predictions, in terms of binary instances and semantic categories. To this end, from the whole framework, which also includes a flow forecasting module, named OFNet, we only take MaskNet, which is a neural network that warps binary instances from the current frame onto the future one. Because MaskNet requires future optical flows to warp instances, we replace OFNet with FLODCAST, by only retaining our flow predictions and discarding depth maps. %, and we use our \textit{FLODCAST + MaskNet} in autoregressive way.
%We compare instance and semantic segmentations generated for future unobserved frames for Cityscapes for every val sequence, both 3 frames and 9 frames ahead (up to about 0.5 sec later), i.e. as done in \cite{luc2018predicting}, respectively named in the literature as short-term and mid-term. 
 %Compared with OFNet, FLODCAST generates for every iteration sequences of future optical flows long $T=3$ rather than just providing one for the next frame, that also contributes to reduce the accumulation error over time (see Tab. \ref{tab:flowresults}, as already pointed out in the section \ref{sec:flowablationstudy}). Basically with respect to OFNet, we iterate FLODCAST autoregressively once for short-term instead of 3, while 3 times instead of 9 to reach the mid-term prediction.

In order to generate future predictions, both instance and semantic segmentations, we follow the same protocol training in \cite{ciamarra2022forecasting}. We first finetune a MaskNet model pretrained on ground truth masks (the MaskNet-Oracle model from \cite{ciamarra2022forecasting}), by feeding future optical flows predicted by FLODCAST. We perform separate trainings to make predictions up to $T+3$ (short-term) and $T+9$ frames ahead (mid-term)\footnote{Note that in the literature there is a slight misalignment when referring to short-term and mid-term, depending on the task. For depth and flow forecasting we refer to short-term mid-term as $T+5$ and $T+10$ and for segmentation forecasting as $T+3$ and $T+9$  respectively.}. We denote these two models as MaskNet-FC. Second, we study how binary instances predicted by MaskNet can be improved. 
Because we employ predicted optical flow to estimate future binary masks, motion mistakes may affect some pixels of the object to be warped. We also believe that some drops in the performance of MaskNet are due to misleading pixels, that are badly labeled as background instead of instance and vice versa.
%Considering that MaskNet-Oracle generates very accurate instances using clean optical flows estimated by FlowNet2\cite{ilg2017flownet}, bad binary masks are produced mainly because of not negligible errors on predicted flows over the objects to be warped, even if the average flow error in the whole scene is small. Moreover, we believe that some drops in the performance of MaskNet on future instance segmentation task is also due to misleading pixels, that are badly labeled as background instead of instance and vice versa. 
This effect is more pronounced when an object appears smaller and its predicted flow is not accurate. 
Inspired by \cite{larrazabal2019anatomical},  we address this issue by introducing a Denoising AutoEncoder network (shortened to DAE) to the output of MaskNet, so to make binary masks cleaner and to make them as much aligned as possible to the ground truth.
%We address this issue by introducing a denoising network trained to mitigate this effect. Inspired by \cite{larrazabal2019anatomical}, we apply a denoising autoencoder model (shortened to DAE) to the output of MaskNet, to provide cleaner binary masks and to make them as much aligned as possible to the ground truth.
%
\begin{figure}[t]
\centering
\includegraphics[width=\linewidth]{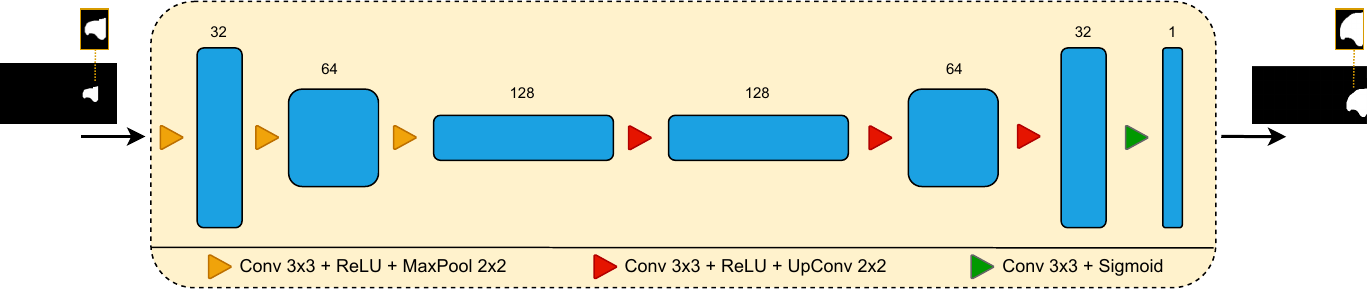}
\caption{Denoising autoencoder (DAE) used to refine the generated future instance segmentation masks. The model is based on a convolutional encoder-decoder structure, where the encoder compresses the input into the latent space and the decoder gradually upsamples the features back to the original image size.}
\label{fig:dae}
\end{figure}
The network, depicted in Fig. \ref{fig:dae}, has an encoder consisting of Conv-ReLU-MaxPool sequences with 32, 64 and 128 filters, and a decoder where Conv-ReLU-UpSample operations are used with 128, 64 and 32 filters. The output is generated after a convolution operation with a single channel, $3 \times 3$ kernel filter and a sigmoid activation function. At inference, outputs are binarized using a 0.5 threshold.

%Since we consider \textit{MaskNet+DAE} as a whole architecture, 
Because MaskNet warps object instances based on optical flows, the generated masks have to be fed to the DAE to get refined. Therefore, we train the DAE, by using autoregressive flows and freezing MaskNet pretrained weights. % so to feed MaskNet outputs in input to the autoencoder. 
Specifically, we train DAE for 3 epochs with a per-pixel MSE loss function with predicted flows up to 3 frames ahead (i.e. $T+3$, short-term).
%Specifically, we first train DAE for 3 epochs with a per-pixel MSE loss function at short-term, through the pretrained weights of the MaskNet short-term model. %At inference time, we attach this pretrained DAE (named \textit{DAE-s}) on top of each MaskNet models and we evaluate performance both at short-term and mid-term predictions. 
We observe that using a Dice loss~\cite{milletari2016v} (already employed to train MaskNet), even in combination with the L2 loss, DAE performs worse than with the MSE function. We believe that is due to the fact that further improvements on instance shapes are not always possible with region-based losses (like Dice loss), instead MSE is more suitable to binarize an instance as a whole image. We continue to finetune the DAE for 3 more epochs using the autoregressive flows predicted up to 9 frames ahead (i.e. $T+9$, mid-term) to adapt the network to less accurate inputs.
%, since instances warped after short-term might have never seen by DAE, thus, they are likely less accurate.
Doing so, we are able to provide a single autoencoder trained to refine instances, which are generated by MaskNet through autoregressive flows predicted up to 9 frames ahead.
%Doing so, we are able to provide a unique autoencoder, which has learned to refine instances generated by MaskNet, which utilizes autoregressive flows predicted up to 9 frames ahead.
%which has been learned to refine instances generated by the MaskNet mid-term model up to 9 time steps after the present. Hence, we evaluate MaskNet+DAE both at short-term and mid-term, by appending this pretrained DAE to the MaskNet mid-term model (named \textit{MaskNet-m}).
%Hence, we evaluate MaskNet+DAE both at short-term and mid-term, by appending DAE to the MaskNet mid-term model. This allows to utilize a single MaskNet model to generate future instance segmentation both at short-term and mid-term. 
Hence, our overall segmentation forecasting architecture, i.e. MaskNet-FC+DAE, is obtained by appending the DAE to the MaskNet mid-term model. This architecture allows to utilize a unique segmentation model to generate future instance segmentation up to 9 frames ahead.

We conduct experiments on the Cityscapes val set, generating future instance and semantic segmentations of 8 different categories of moving objects, both 3 frames and 9 frames ahead (up to about 0.5 sec later) as done in \cite{luc2018predicting}, respectively referred to in the literature as short-term and mid-term. We use the mAP and mAP50 metrics for instance segmentation, and mIoU (mean IoU) for semantic segmentation.
%The metric mAP50 is the mean average precision that counts an instance as correct if matching its groundtruth for at least 50\% of intersection-over-union (IoU). Instead, mAP is obtained by averaging Average Prevision at 10 equally spaced IoU thresholds from 50\% to 95\%. 
We show our quantitative results in Table \ref{table:segmentationresults}.
\begin{table}[t]
\centering
%\fontsize{6pt}{6pt}\selectfont
%\caption{Results for future instance segmentation (AP and AP50) and future semantic segmentation (IoU) of moving objects on Cityscapes val set. "MaskNet-m" is the pretrained MaskNet mid-term model, "DAE-s" is the autoencoder trained using outputs generated by the MaskNet short-term model, while "DAE" is obtained by finetuning "DAE-s" with MaskNet-m. Best results are in bold, second best are underlined.}\label{table:segmentationresults}
\caption{Future instance segmentation (AP and AP50) and future semantic segmentation (IoU) of moving objects on the Cityscapes val set. Best results in bold, second best underlined.}\label{table:segmentationresults}
\resizebox{.9\linewidth}{!}{
\begin{tabular}{@{}ccccccc@{}}
\toprule
Method & \multicolumn{3}{c}{Short term (T+3)} & \multicolumn{3}{c}{Mid term (T+9)} \\ 
%Method & \multicolumn{3}{c}{$t+3$} & \multicolumn{3}{|c}{$t+9$} \\
 & ~~AP~~ & ~~AP50~~ & ~~IoU~~ & ~~AP~~ & ~~AP50~~ & ~~IoU~~ \\ \midrule
Mask RCNN oracle & 34.6 & 57.4 & 73.8 & 34.6 & 57.4 & 73.8 \\ \midrule
MaskNet-Oracle \cite{ciamarra2022forecasting} & 24.8 & 47.2 & 69.6 & 16.5 & 35.2 & 61.4\\ \midrule
Copy-last segm. \cite{luc2018predicting} & 10.1 & 24.1 & 45.7 & 1.8 & 6.6 & 29.1 \\
Optical-flow shift \cite{luc2018predicting} & 16.0 & 37.0 & 56.7 & 2.9 & 9.7 & 36.7 \\
Optical-flow warp \cite{luc2018predicting} & 16.5 & 36.8 & 58.8 & 4.1 & 11.1 & 41.4 \\
Mask H2F \cite{luc2018predicting} & 11.8 & 25.5 & 46.2 & 5.1 & 14.2 & 30.5 \\
F2F \cite{luc2018predicting} & \underline{19.4} & \underline{39.9} & 61.2 & \textbf{7.7} & 19.4 & 41.2 \\
MaskNet \cite{ciamarra2022forecasting} & \textbf{19.5} & \textbf{40.5} & \textbf{65.9} & 6.4 & 18.4 & 45.5 \\ \midrule
%(Ours) & & & & & & \\
MaskNet-FC & 18.1 & 37.8 & 65.4 & 6.7 & \underline{18.9} & \underline{48.4} \\
%FLODCAST + MaskNet w DAE-s & 17.9 & 37.5 & 65.4 & 7.0 & \underline{19.8} & \underline{49.1}\\
%FLODCAST + MaskNet-m w DAE & 18.3 & 39.0 & \underline{65.7} & \underline{7.1} & \textbf{20.7} & \textbf{49.2}\\ \bottomrule
MaskNet-FC+DAE (Ours) & 18.3 & 39.0 & \underline{65.7} & \underline{7.1} & \textbf{20.7} & \textbf{49.2}\\ \bottomrule
\end{tabular}}
\end{table}

We report segmentation results achieved by MaskNet \cite{ciamarra2022forecasting}, using flows predicted by our FLODCAST, also considering the denoising autoebcoder (DAE), proposed to refine warped masks. We compare our results with the original flow-based approach MaskNet \cite{ciamarra2022forecasting}. We also report the oracle reference, where a Mask RCNN \cite{he2017mask} is used directly on future frames, as well as MaskNet-Oracle whose model is our upper bound flow-based approach since segmentations are warped using ground truth flows. Moreover, we listed the performances of 4 simple baselines and the commonly used F2F approach \cite{luc2018predicting}.

\begin{figure}[t]
\centering
\resizebox{\linewidth}{!}{\begin{tabular}{cccc}
\midrule
\noalign{\smallskip} \multicolumn{2}{c}{Short-term (T+3)} & \multicolumn{2}{c}{Mid-term (T+9)} \\
\noalign{\smallskip} Mask R-CNN Oracle & Our predictions & Mask R-CNN Oracle & Our predictions \\ \midrule
\noalign{\smallskip} \includegraphics[width=.3\linewidth]{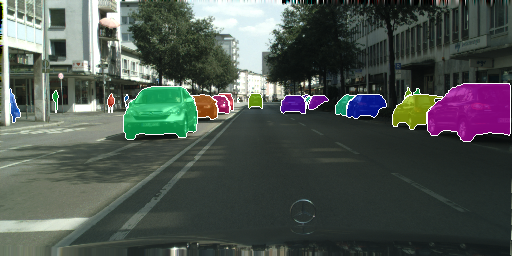} & \includegraphics[width=.3\linewidth]{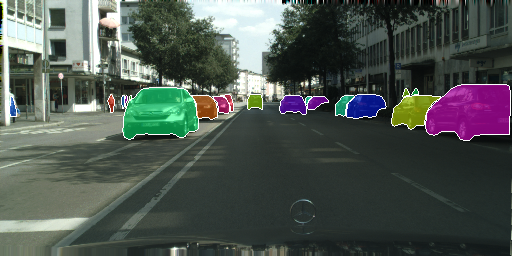} & \includegraphics[width=.3\linewidth]{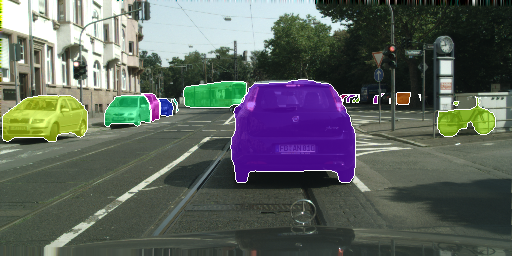} & \includegraphics[width=.3\linewidth]{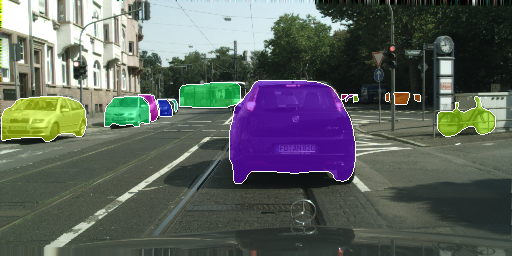} \\
\noalign{\smallskip} \includegraphics[width=.3\linewidth]{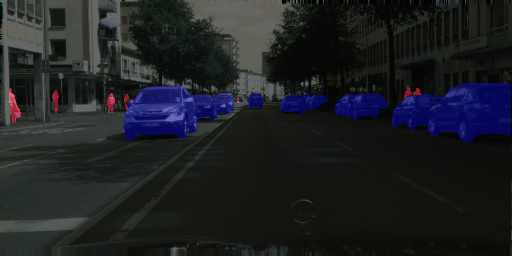} & \includegraphics[width=.3\linewidth]{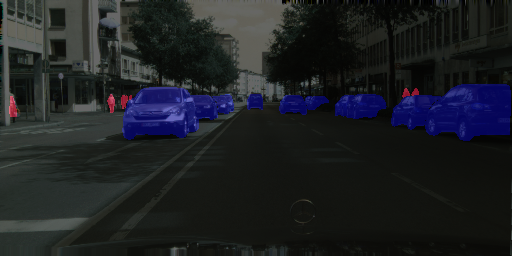} & \includegraphics[width=.3\linewidth]{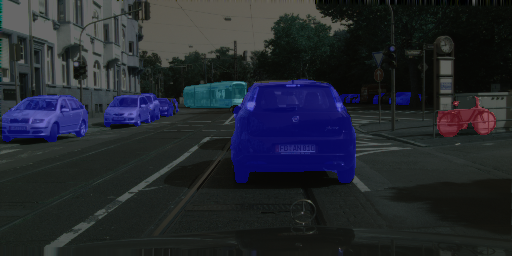} & \includegraphics[width=.3\linewidth]{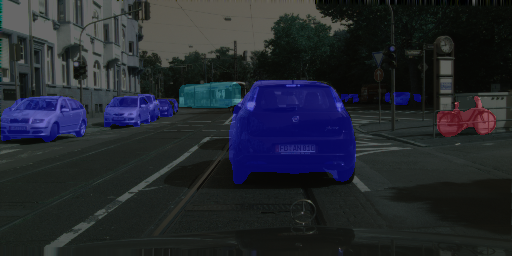} \\
\includegraphics[width=.3\linewidth]{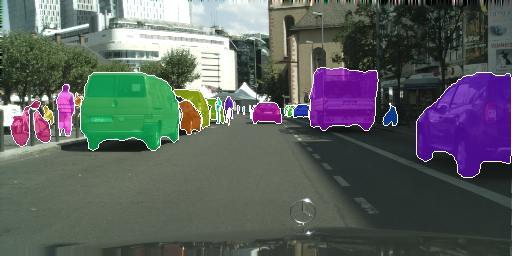}  & \includegraphics[width=.3\linewidth]{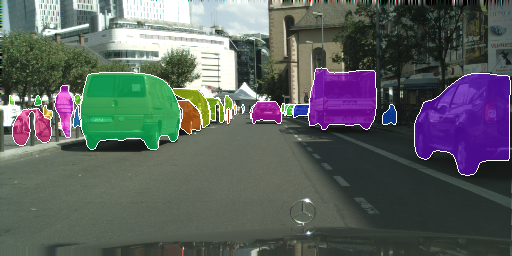} & \includegraphics[width=.3\linewidth]{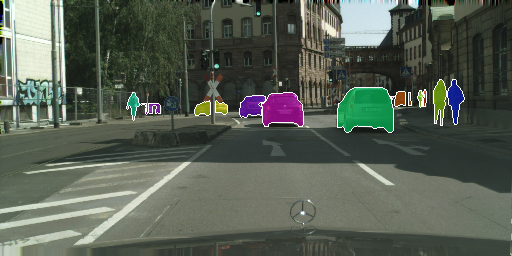} & \includegraphics[width=.3\linewidth]{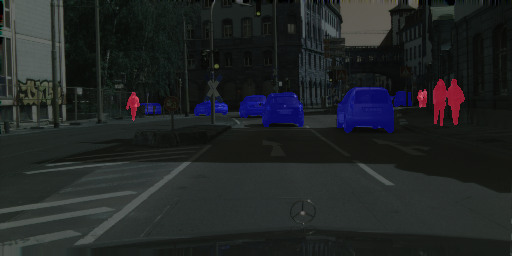} \\
\includegraphics[width=.3\linewidth]{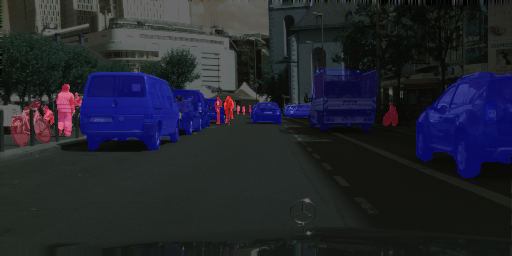}  & \includegraphics[width=.3\linewidth]{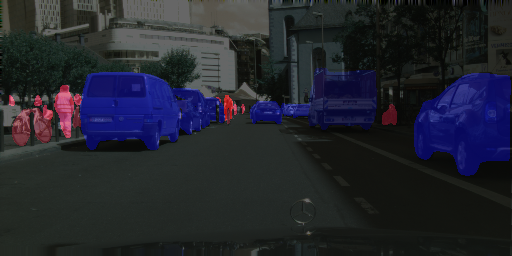} & \includegraphics[width=.3\linewidth]{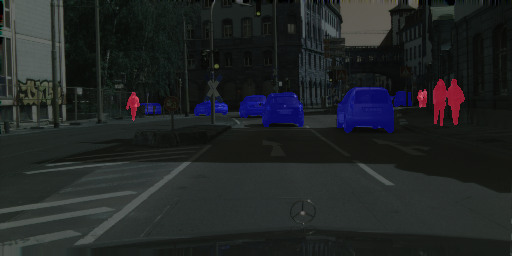} & \includegraphics[width=.3\linewidth]{images/masknet/samples/t+9/003/pred_segm_frankfurt_000001_008688_leftImg8bit.png} \\ \bottomrule
\end{tabular}}
\caption{Qualitative results of future instance and semantic segmentation predictions on the Cityscapes val set both at short-term and mid-term generated by MaskNet-FC+DAE.}\label{fig:masknetresults}
\end{figure}

We found that MaskNet, using flows predicted by FLODCAST, improves at mid-term, getting +0.5\% and +2.9\%, respectively for instance and semantic segmentations compared to the original formulation of \cite{ciamarra2022forecasting}. Meanwhile, we observe a negligible drop at short-term, since FLODCAST generates more accurate flows after the first iteration. Because the segmentation performance typically degrades over the time, we pay attention to the impact of appending our DAE at the end of MaskNet to enhance instance and semantic results mainly at mid-term (i.e. 9 frames ahead, 0.5 sec), which is a more challenging scenario than the short-term one.
%We found that appending our autoencoder DAE trained at short-term (i.e. \textit{DAE-s}) but evaluated at mid-term, improves the instance and semantic segmentation results, by +0.9\% AP50 and +0.7\% IoU respectively, with respect to FLODCAST+MaskNet, and consequently exceeds OFNet+MaskNet. 
When the DAE is trained to refine instance masks up to mid-term we report a considerable improvement against the F2F approach with a gain of +1.3\% in AP50 and +8\% in IoU.
%In additional, simply training some convolutional layers at the end of MaskNet gets still performance at mid-term, obtaining 19.8\% and 49.1\% in terms of instance and semantic segmentations, thus FLODCAST+MaskNet using DAE definitely beats the F2F approach by a margin of 1.0\% in mAP50 and +7.7\% in mIoU respectively. 
Some qualitative results of future instance and semantic segmentation are shown in Fig. \ref{fig:masknetresults}.

\begin{figure}[t]
\centering
\resizebox{\linewidth}{!}{
\begin{tabular}{cccc}
\toprule
& \multicolumn{3}{l}{Short-term (T+3)} \\
\noalign{\smallskip} RGB & Mask R-CNN Oracle & MaskNet \cite{ciamarra2022forecasting} & Our predictions \\ \midrule
\noalign{\smallskip} \includegraphics[width=.5\linewidth]{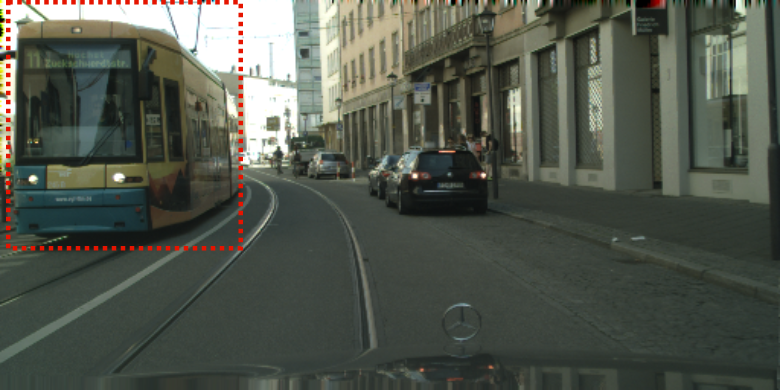} & \includegraphics[width=.25\linewidth]{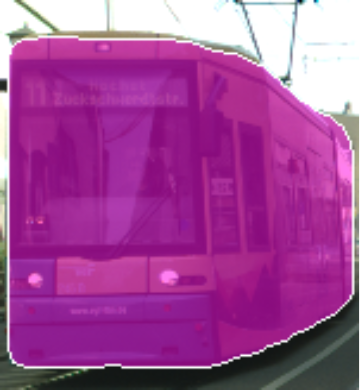} & \includegraphics[width=.25\linewidth]{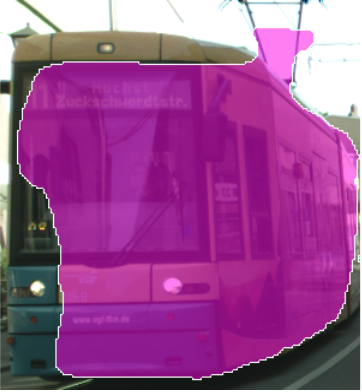} & \includegraphics[width=.25\linewidth]{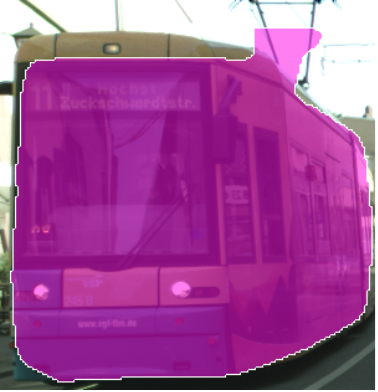} \\
\noalign{\smallskip} \includegraphics[width=.5\linewidth]{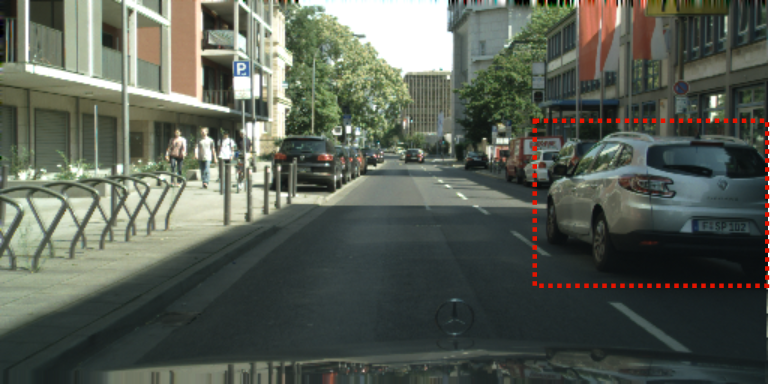} & \includegraphics[width=.25\linewidth]{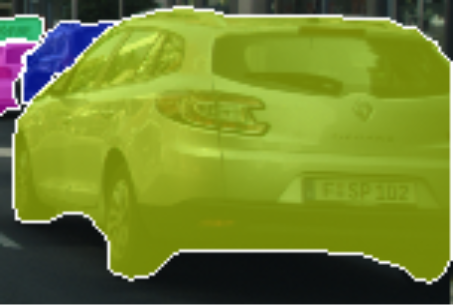} & \includegraphics[width=.25\linewidth]{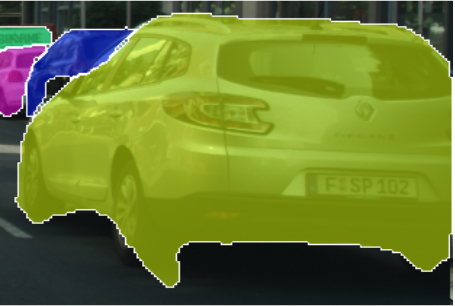} & \includegraphics[width=.25\linewidth]{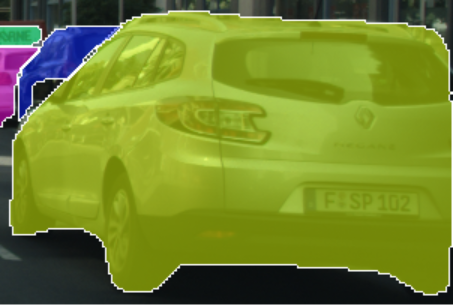} \\ \bottomrule
\end{tabular}}
\caption{Some qualitative results of future instance segmentation predictions on the Cityscapes val set 3 frame ahead (short-term).}\label{fig:masknetinstresults3}
\end{figure}

\begin{figure}[t]
\centering
\resizebox{\linewidth}{!}{
\begin{tabular}{cccc}
\toprule
%\noalign{\smallskip} \multicolumn{4}{c}{Mid-term $t+9$} \\
& \multicolumn{3}{l}{Mid-term (T+9)} \\
\noalign{\smallskip} RGB & Mask R-CNN Oracle & MaskNet \cite{ciamarra2022forecasting} & Our predictions \\ \midrule
\includegraphics[width=.5\linewidth]{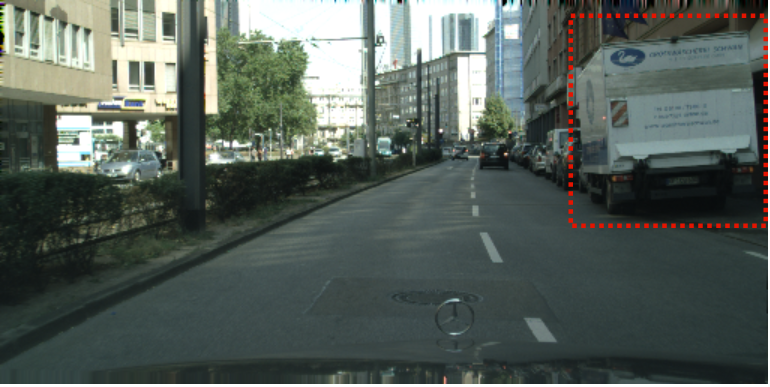} & \includegraphics[width=.25\linewidth]{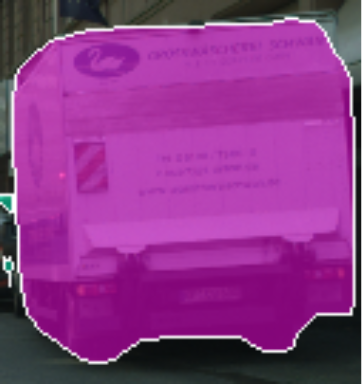} & \includegraphics[width=.25\linewidth]{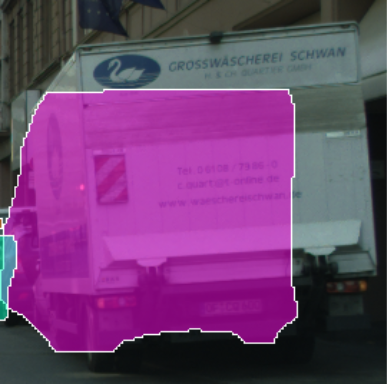} & \includegraphics[width=.25\linewidth]{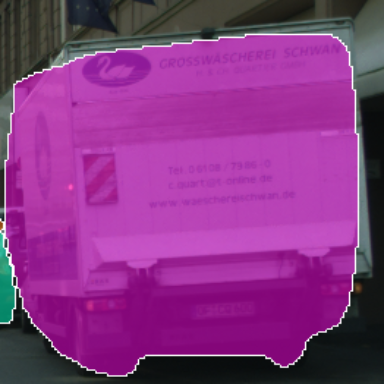} \\
\includegraphics[width=.5\linewidth]{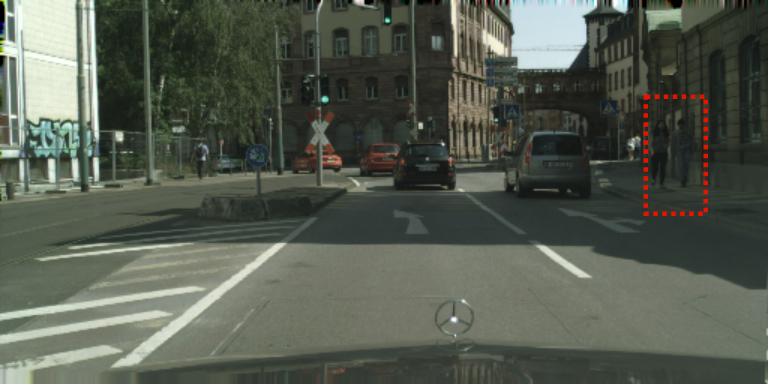} & \includegraphics[width=.13\linewidth]{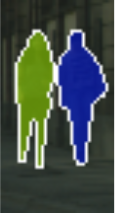} & \includegraphics[width=.13\linewidth]{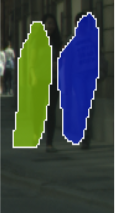} & \includegraphics[width=.13\linewidth]{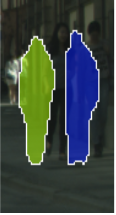} \\ 
\includegraphics[width=.5\linewidth]{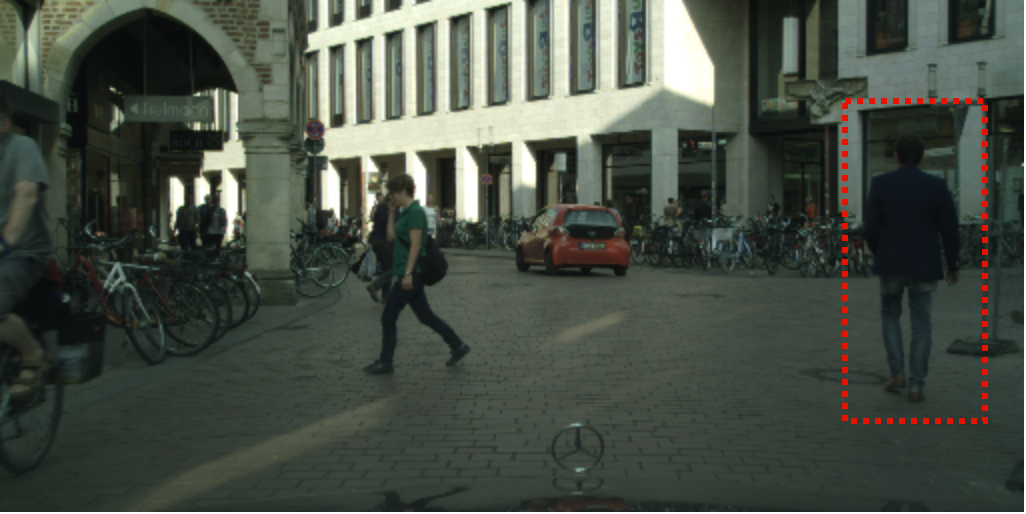} & \includegraphics[width=.13\linewidth]{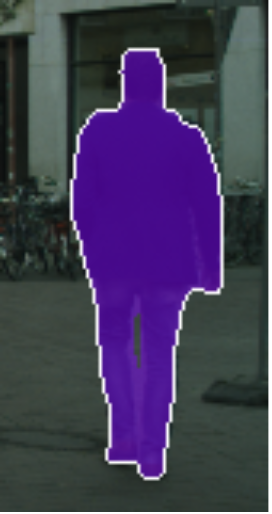} & \includegraphics[width=.13\linewidth]{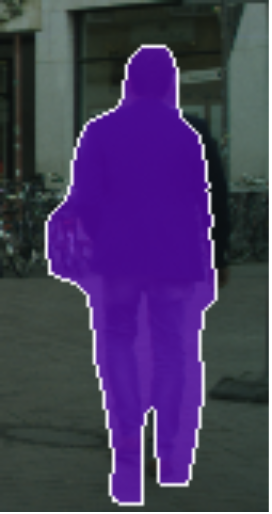} & \includegraphics[width=.13\linewidth]{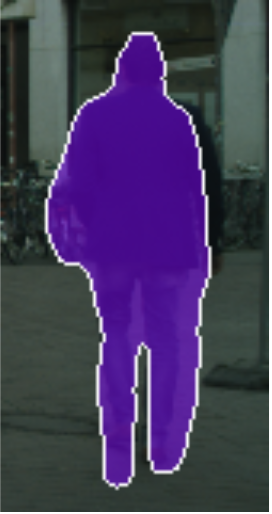} \\ \bottomrule
\end{tabular}}
\caption{Some qualitative results of future instance segmentation predictions on the Cityscapes val set 9 frames ahead (mid-term).}\label{fig:masknetinstresults9}
\end{figure}

We additionally provide some qualitative results in terms of instance segmentations predicted, by using FLODCAST and MaskNet-FC+DAE, in comparison with the previous framework, i.e. OFNet and MaskNet. We show enhancements on different objects and shapes predicted both at short-term (Fig. \ref{fig:masknetinstresults3}) and mid-term (Fig. \ref{fig:masknetinstresults9}), such as the big shapes (like trams and trucks) as well as some details (like car wheels and pedestrians on the ground).
%To this end, in Fig \ref{fig:masknetinstresults} we show enhancements on different objects and shapes predicted both at short-term and mid-term (e.g. see the wheel of the car or pedestrians on the pavement).

%

\section{Conclusions}

In this work, we proposed FLODCAST, a novel multimodal and multitask network able to jointly forecast future optical flows and depth maps using a recurrent architecture. Differently from prior work, we forecast both modalities for multiple future frames at a time, allowing decision-making systems to reason at any time instant and yielding state-of-the-art results up to 10 frames ahead on the challenging Cityscapes dataset.
We demonstrated the superiority of exploiting both optical flow and depth as input data against single-modality models, showing that leveraging both modalities in input can improve the forecasting capabilities for both flow and depth maps, especially at farther time horizons.
%This motivates our design choices which follow the intuition, based on the fact that displacements of objects moving at different resolutions highly affect predictions, either optical flow and depth, as soon as one data source is excluded.
We also demonstrated that FLODCAST can be applied on the downstream task of segmentation forecasting, relying on a mask-warping architecture, improved with a refining instance model that boosts mid-range predictions.

Further research will be considered for future developments, which include the usage of a transformer architecture to boost our multitasking model. Other lines of research may also include more performing mask-level segmentation models to be trained end-to-end with a flow forecasting architecture, in order to directly perform the task for multiple frames at a time, in the same sense FLODCAST was designed.

\bibliography{ref}

\begin{wrapfigure}{l}{25mm} 
\includegraphics[width=1.1in,height=1.5in,clip,keepaspectratio]{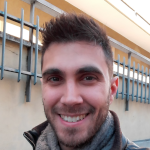}
\end{wrapfigure}
\par \textbf{Andrea Ciamarra} achieved a master's degree in computer engineering cum laude in 2019 from the University of Florence, with a thesis related to vehicle motion predictions through optical flow. Currently, he is a Ph.D student at Media Integration and Communication Center and working on forecasting tasks for behavior anticipation in automotive.
\\
\begin{wrapfigure}{l}{25mm} 
\includegraphics[width=1.1in,height=1.5in,clip,keepaspectratio]{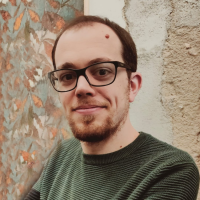}
\end{wrapfigure}
\par \textbf{Federico Becattini} is a Tenure-Track Assistant Professor at the University of Siena. His research focuses on computer vision and memory-based learning. He organized tutorials and workshops at ICPR2020, ICIAP2020, ACMMM2022, ICPR2022, ECCV2022. He has co-authored more than 40 papers. He is Associate Editor of the International Journal of Multimedia Information Retrieval (IJMIR).
\\
\begin{wrapfigure}{l}{25mm} 
\includegraphics[width=1.1in,height=1.5in,clip,keepaspectratio]{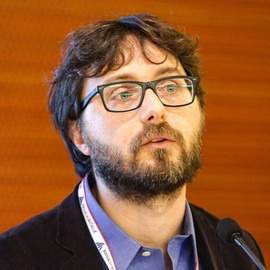}
\end{wrapfigure}
\par \textbf{Lorenzo Seidenari} is an Associate Professor at the University of Florence. His research focuses on deep learning for object and action recognition. He is an ELLIS scholar. He was a visiting scholar at the University of Michigan in 2013. He gave a tutorial at ICPR 2012 on image categorization. He is author of 16 journal papers and more than 40 peer-reviewed conference papers.
\\
\begin{wrapfigure}{l}{25mm} 
\includegraphics[width=1.1in,height=1.5in,clip,keepaspectratio]{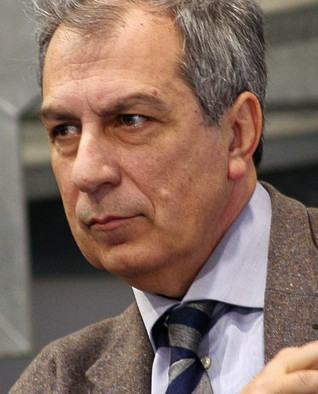}
\end{wrapfigure}
\par \textbf{Alberto Del Bimbo} is an Emeritus Professor and was director of the Media Integration and Communication Center at University of Florence. His interests are multimedia and computer vision. He was General Co-Chair of ACMMM2010, ECCV2012, ICPR2020. ACM nominated him Distinguished Scientist and he received the SIGMM Technical Achievement Award for Outstanding Technical Contributions to Multimedia Computing, Communications and Applications.

\end{document}